\documentclass[runningheads]{llncs}
\usepackage[T1]{fontenc}
\usepackage{newtxtext}       %
\usepackage[varvw]{newtxmath}       
\usepackage{xcolor}

\usepackage{amsmath}
\usepackage{todonotes}
\usepackage{amsfonts}
\usepackage{mathrsfs}
\usepackage{pifont}
\usepackage{makecell}
\usepackage{booktabs}
\usepackage{array}
\usepackage{multirow}
\usepackage{algorithm}
\usepackage{algpseudocode}
\usepackage{lscape}
\usepackage{adjustbox}
\usepackage{makecell}
\usepackage{tabularx}
\usepackage{orcidlink}
\usepackage[table]{xcolor}
\usepackage{orcidlink}
\usepackage{orcidlink}
\renewcommand{\orcidID}[1]{\,\orcidlink{#1}}
\usepackage{bm}

\usepackage{graphicx}
\begin{document}
\title{Process-Aware Event Log Augmentation for Process Prediction under Labeled Data Scarcity}
\titlerunning{Process-Aware Event Log Augmentation for PPM under Labeled Data Scarcity}
%
%
\author{Sjoerd van Straten\inst{1}\orcidID{0009-0002-5772-9073} \and
Alessandro Padella\inst{2}\orcidID{0009-0008-6399-9364} \and
Marwan Hassani\inst{3,1}\orcidID{0000-0002-4027-4351}}
\authorrunning{S. van Straten et al.}
%
\institute{Eindhoven University of Technology, Eindhoven, The Netherlands\\
\email{h.a.j.v.straten@tue.nl} \and
University of Padua, Padua, Italy\\
\email{alessandro.padella@unipd.it} \and
University of Wuppertal, Wuppertal, Germany\\
\email{hassani@uni-wuppertal.de}}
\maketitle              

\begin{abstract}
Predictive Process Monitoring (PPM) forecasts future events of ongoing business processes based on event logs. However, deep learning PPM approaches are often limited by labeled data scarcity and the low variability of real-world logs. To address this, we introduce SiamSA-PPM, a self-supervised framework combining Siamese learning with statistically grounded data augmentation. It employs novel transformation methods that leverage control-flow semantics and frequent behavioral patterns to generate realistic, semantically valid trace variants, learning generalizable prefix representations without labeled supervision. Extensive experiments on real-world event logs with a wide range of variability, incorporating exhaustive hyperparameter tuning, demonstrate the value of this process-aware augmentation. Compared to state-of-the-art baselines, SiamSA-PPM achieves competitive or superior predictive performance in both next activity and final outcome prediction tasks. An open source implementation of our method is publicly available under: \url{https://github.com/SvStraten/SiamSA-PPM-IJCIS}.
\end{abstract}


\section{Introduction}
\label{sec:introduction}
Due to the increasing digitization of business operations, Predictive Process Monitoring (PPM) has become a critical area of research within process mining~\cite{ceravolo2024predictive}. Two of its core tasks, \textit{next activity prediction} and \textit{final outcome prediction}, have received substantial attention due to their practical relevance in optimizing workflows, reducing delays and improving resource utilization \cite{Verbeek2025}. While earlier works utilized symbolic models such as Petri nets and transition systems, the advancements in deep learning, especially models such as Long Short-Term Memory (LSTM)~\cite{tax2017predictive} and Attention-based Graph Transformers~\cite{vanderaa2024pgtnet}, have significantly improved the predictive performance of such models. However, a persistent challenge in deep learning and inherently in its application to PPM is the limited size of available labeled data \cite{ceravolo2024predictive,vanstraten2026dtaia}. Despite their high complexity, business processes are typically logged with limited trace variation and often result in datasets containing redundant or highly similar samples. These limitations affect the generalizability of predictive models, especially in scenarios with infrequent behaviors or class imbalance. 

\begin{figure}[t]
    \centering
    \includegraphics[width=1\linewidth]{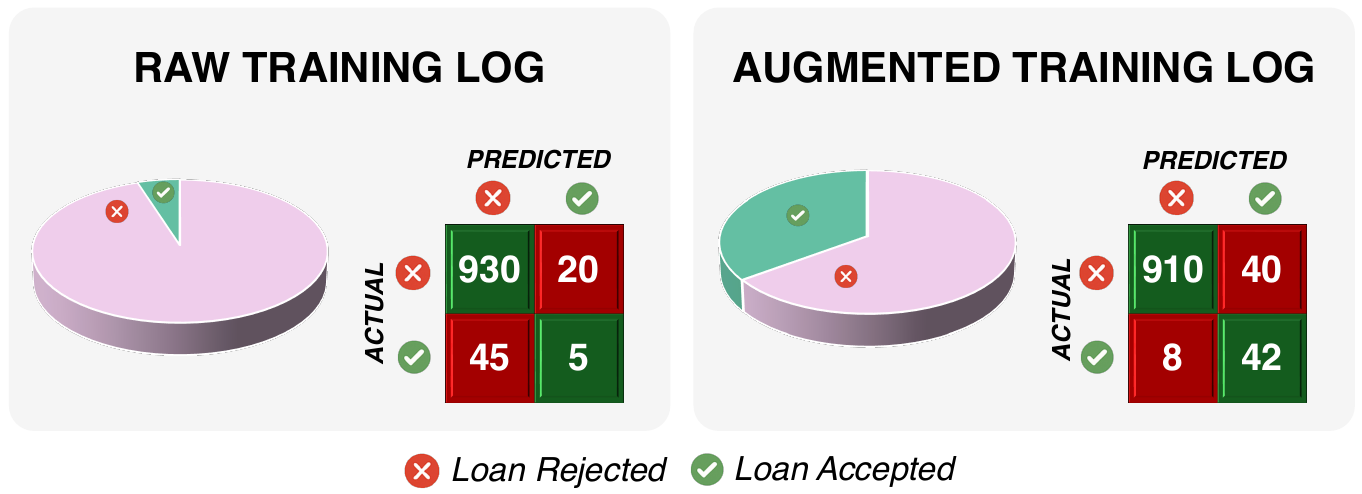}
    \vspace{-1.5em}
    \caption{The impact of class imbalance and data augmentation on predictive models using a loan application scenario. Training on a raw, severely imbalanced log (left) leads to biased predictions. The model achieves a high overall accuracy by favoring the majority class but fails to detect the critical minority class. By augmenting the training log to balance the distributions (right), the model can successfully learn the patterns of the rare cases, drastically improving performance on the minority class.}
    \label{fig:introduction_augment}
\end{figure}

A practical example is a loan application process in which 99\% of applications are accepted and only 1\% are rejected. The corresponding event log is therefore severely class‑imbalanced. In such a setting, both standard machine learning models and data‑hungry deep‑learning predictors tend to produce biased predictions that favor the majority class, thereby failing to reliably detect the critical minority class that lenders need to identify manually (cf. Fig.~\ref{fig:introduction_augment}). In this scenario, balancing the event log by removing a selected number of acceptance traces is not feasible, as this would drastically reduce the available data to only 2\% of the original volume. 

\begin{figure}[t]
    \centering
    \includegraphics[width=0.7\linewidth]{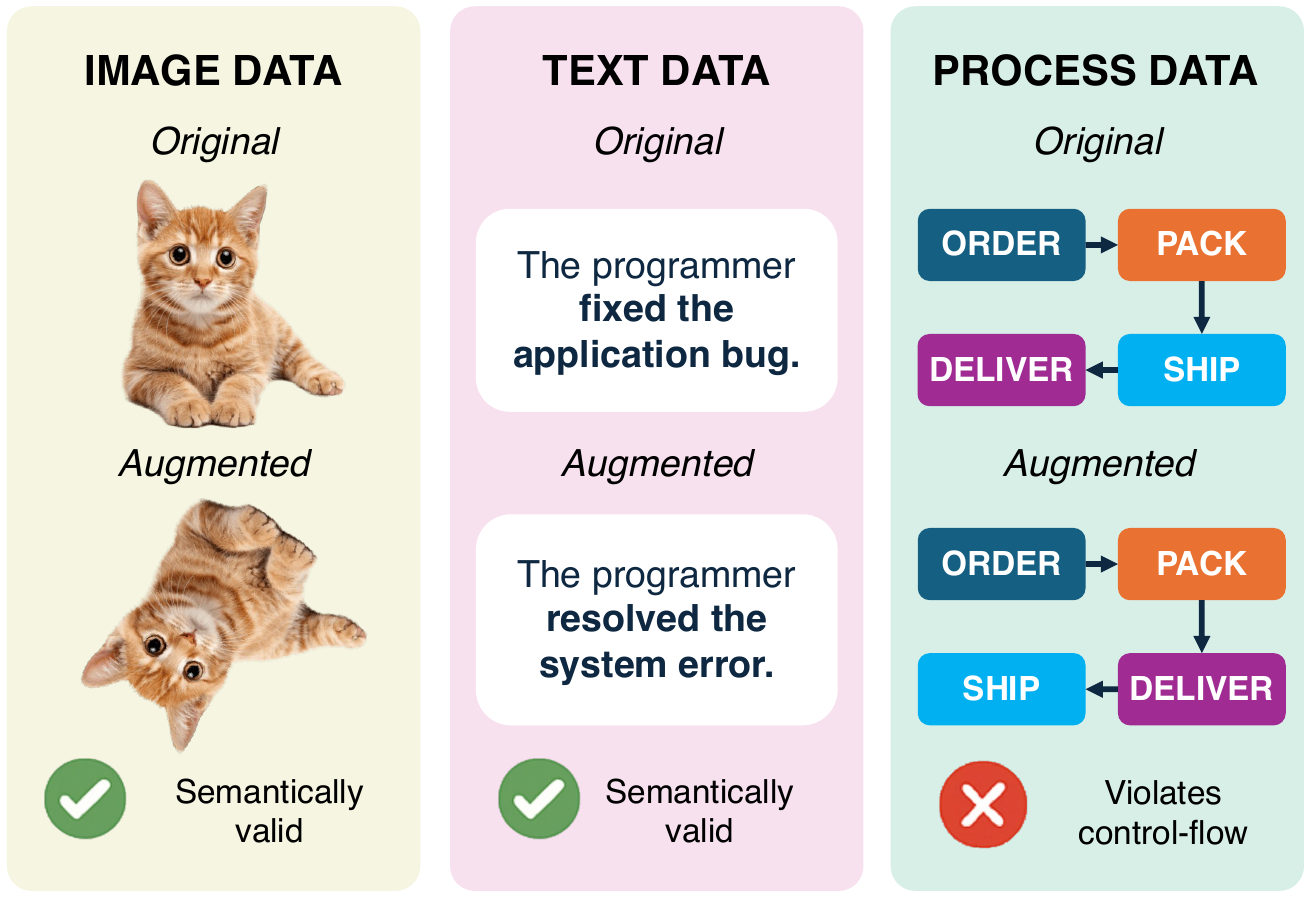}
    \caption{Random augmentation strategies can preserve semantics in image and text domains, where structural flexibility exists. However, in process data, such augmentations risk violating control-flow dependencies for instance, producing logically inconsistent or non-executable sequences.}
    \label{fig:introduction_image}
\end{figure}

In machine learning, a common remedy for such constraints is \textit{data augmentation}: the process of artificially generating additional training data to enhance diversity and robustness. Data augmentation has been extensively explored in domains such as Computer Vision (CV) and Natural Language Processing (NLP)~\cite{wang2024comprehensive}. In these fields, augmentations are informed by domain knowledge to preserve the semantic integrity of the data. For example, visual enhancements such as flipping or cropping retain object identity, while textual methods such as synonym replacement, word deletion or insertion aim to maintain grammatical and contextual coherence~\cite{wei2019eda}. The core principle in the data augmentation domain is that augmented data should be \textit{semantically valid} and \textit{task-relevant}~\cite{miao2026augmentation}. However, no existing approach matches these two requirements: \begin{enumerate}
    \item Data augmentation methods in literature are grounded in the assumption that samples come from the same distribution and that they are independent. However, this assumption does not hold in process mining, where events in traces are inherently dependent~\cite{AalstBook}. As a result, generic augmentation methods are not task-relevant for process mining, which justifies the need for tailored data augmentation methods.

    \item Existing prior work on data augmentation specifically tailored for event logs in PPM investigates model-agnostic augmentation based mainly on random insertions, deletions, or replacements~\cite{kappel2023augmentation}. However, such random transformations run the risk of violating process semantics, for instance by generating an event such as delivering an order before it has been shipped (cf.\ Fig.~\ref{fig:introduction_image}).

\end{enumerate}

Our approach bridges this gap. It introduces statistically grounded augmentation techniques that leverage frequent control-flow patterns to generate realistic, semantically valid trace variants, thereby aligning augmentation with actual process dynamics (cf. Section~\ref{sec:related_work}). We use the process in Fig.~\ref{fig:introduction_image} $(Order \rightarrow Pack \rightarrow Ship \rightarrow Deliver)$ as a running example throughout the paper to illustrate how random augmentations can violate semantics, while our statistically grounded method is designed to ensure, on the one hand, adherence to constraints inferred in a data‑driven manner, and on the other hand, task relevance, as it is explicitly tailored to and designed for process mining. However, this benefit does not come in isolation. Event logs record the observed behavior of a given process; when the log is scarce, there is a risk that not all possible behaviors are represented, since certain traces may, in theory, occur but have not yet been recorded. By performing event‑log augmentation that is both semantically valid and task‑relevant, we also enrich the log in terms of contained information, artificially generating traces that could plausibly occur but remain unobserved due to the limited availability of data.

In addition to augmentation, we also explore the underutilized potential of \textit{self-supervised learning} (SSL) in PPM. While most existing approaches rely on supervised learning, this requires labeled data (e.g. next activity or final outcome annotations) for each trace. Practically, such an amount of labels may not be sufficient to train a model. In fact, some event logs may be imbalanced, resulting in for example an insufficient number of examples for a given outcome, e.g. a given machine breaks down. This issue also applies to the next activity prediction task, where the occurrence of certain activities may be rare, leading to a very limited number of events with the corresponding label. SSL addresses this by learning from the data itself without requiring manual labels. It constructs surrogate tasks, such as aligning different augmented views of the same input to learn robust task-agnostic representations. Although SSL has seen great success in the computer vision domain \cite{chen2020simclr,grill2020byol,chen2021simsiam}, it is underexplored in PPM. 

In this work, we adopt a Siamese learning strategy based on Bootstrap Your Own Latent (BYOL) \cite{grill2020byol}, where augmented trace prefixes are encoded into similar latent representations. This setup avoids the need for negative samples or large batch sizes \cite{grill2020byol}, making it well-suited for structured, small-scale event logs. Combined with our statistically grounded augmentations, this enables high-quality representations for downstream tasks, even with limited labeled data. The contributions of this paper include:

\begin{itemize}
    \item Introducing three novel, statistically grounded trace transformation methods that preserve behavioral semantics and process structure.
    \item Introducing a self-supervised Siamese framework that learns robust embeddings of process prefixes using augmented trace pairs.
    \item Conducting extensive experiments on eight real-world case studies, demonstrating the performance of our approach in both next activity prediction and final outcome prediction.
    \item Comparing our framework against four state-of-the-art baselines, analyzing predictive performance and computational efficiency.
\end{itemize}

This paper extends, but fundamentally differs from our paper in \cite{van2025leveraging}. Among others, this paper extends \cite{van2025leveraging} by i) conducting an extended in-depth evaluation distinguishing between target datasets (smaller or low-variability event logs, where augmentation is expected to matter most) and highly variable datasets (cf. Section~\ref{subsec:datasets}), including a non-deterministic analysis; ii) performing extensive hyperparameter tuning across all evaluated baselines and datasets to ensure a fair comparison; iii) adding an elaborate discussion and future work section detailing the practical trade-offs and opportunities of our approach; and iv) providing an extended literature review with a comprehensive taxonomy to clarify the research gap.

The remainder of this paper is organized as follows. Section \ref{sec:related_work} discusses related work in data augmentation, self-supervised learning and PPM. Section \ref{sec:problem_definition} formally defines the problem setting. Section \ref{sec:method} presents our proposed method, Siamese learning with Statistical Augmentation for PPM (SiamSA-PPM), including the augmentation strategy and the SSL framework. Next, Section \ref{sec:experiments} reports on our experimental setup, after which Section~\ref{sec:results} shows our results with an associated ablation study. Finally, Section \ref{sec:conclusion} concludes with a discussion and directions for future work.
\section{Related Work}
\label{sec:related_work}

\subsection{Data Augmentation Techniques}\label{subsec:data_augmentation}
Data augmentation is a proven strategy to increase training data and improve model generalization. In the NLP domain, well-known fundamental methods like EDA~\cite{wei2019eda} introduce random insertion, deletion and replacement of words. In CV, simple pixel-level and mixing techniques (e.g. ReMix~\cite{chou2020remix} and SmoothMix~\cite{lee2020smoothmix}) enhance robustness. Recently, both fields have moved beyond naive sample generation toward highly principled data augmentation strategies. In CV, attribution-guided methods focus on semantically informative regions~\cite{Mi2025CVPR}, while lightweight statistical injection tackles feature skew in federated settings~\cite{Yan2025CVPR}. Diffusion-based augmentation and hard sample synthesis further push performance in data-scarce regimes~\cite{Zhang2025CVPR,Das2025CVPR,Rahimi2025NeurIPS,Kim2025NeurIPS}. Similar trends emerge in NLP, where diversity-oriented frameworks fine-tune LLMs to broaden distributional coverage~\cite{Wang2025ACL} and dynamic strategies diagnose model-specific weaknesses to synthesize targeted training data~\cite{Ji2025EMNLP}.

In stark contrast to the advanced, targeted strategies in NLP and CV, data augmentation in PPM remains largely underexplored. The work in~\cite{kappel2023augmentation} introduced the first comprehensive, model-agnostic augmentation framework for event logs. Their method applies a set of timestamp-preserving transformations, including random insertion, deletion, replacement and swap (derived from~\cite{wei2019eda}), while also introducing loop augmentation and the swapping of events with the same timestamp. Although these transformations increase variability, a key limitation is that they are applied at random, without considering the underlying process structure. This can result in unrealistic behavior that contradicts the actual control-flow semantics of the process. To address this lack of process-awareness, very recent work explores the use of LLMs to generate valid, previously unseen process executions via complex prompt engineering~\cite{KAPPEL2026102717}. While LLM-based augmentation generates more meaningful traces than random baselines, it introduces significant computational overhead, relies heavily on the manual injection of domain knowledge and suffers from prompt brittleness~\cite{KAPPEL2026102717}. Beyond event logs, data augmentation has also been applied to process information extraction from natural language text~\cite{neuberger2026repeat}.

As such, existing PPM augmentation methods either fail to guarantee that new traces are likely within the process behavior or demand expensive, highly manual generative deep learning setups. This gap motivates the development of statistically grounded transformations, based on the frequent paths or XOR-splits mined directly from the original training data. Such augmentations are designed to maintain control-flow integrity while introducing meaningful variability to support generalization, avoiding the overhead and hallucinations of generative models.

\subsection{Self-Supervised Learning Approaches}\label{subsec:ssl_approaches}
SSL has emerged as a powerful tool to learn useful representations from unlabeled data. Instead of relying on manual annotations, SSL defines pretext tasks that use the data itself to generate training input. This has proven particularly effective in domains where labeled data is scarce or expensive to obtain. A dominant class of SSL methods uses contrastive learning techniques. SimCLR \cite{chen2020simclr} generates two augmented views of the same input and trains the model to bring these closer in the embedding space while pushing apart other (negative) samples. While effective, contrastive methods require careful design of negative pairs, large batch sizes and heavy data augmentation. This makes them difficult to apply to structured, small-scale domains such as PPM. To overcome these limitations, Siamese learning frameworks like BYOL \cite{grill2020byol} and SimSiam \cite{chen2021simsiam} have gained popularity. These models use two identical networks to process different augmented versions of the same input, learning to align their latent representations. Unlike contrastive methods, they do not require negative pairs or large batch sizes. BYOL introduces a momentum-updated target network, while SimSiam claims to achieve similar performance with simpler architectures and a stop-gradient mechanism \cite{chen2021simsiam}. Both frameworks have shown strong results in learning high-quality embeddings across NLP and computer vision tasks. Recently, DESPT \cite{SAC25} advanced contrastive learning and correlation in time series to improve the prediction of traffic flow on network sensors without any previously seen training data.

SSL enables representation learning from unlabeled data, which can benefit PPM tasks such as next activity prediction and final outcome prediction. Pre-training with SSL can also help when labeled traces are limited or imbalanced, which is often the case, especially for final outcome prediction. Moreover, SSL encourages the model to learn structural and semantic properties of the data, rather than overfitting to specific task labels. However, the quality of learned representations is clearly dependent on the diversity and the relevance of the augmentations used to create positive pairs. This makes statistically grounded data augmentation not just beneficial, but essential for effective representation learning in PPM.

\subsection{Offline PPM Approaches}\label{subsec:ppm_approaches}
PPM aims to forecast the future behavior of ongoing process instances such as the next activity, remaining time or final outcome, based on event log data. While early approaches relied on formal models (e.g. Petri nets and transition systems), deep learning models now dominate due to their ability to learn complex patterns~\cite{ceravolo2024predictive}. In particular, models such as LSTMs \cite{tax2017predictive} and Convolutional Neural Networks (CNNs) \cite{dimauro2019activity,pasquadibisceglie2019cnn} have been widely used for PPM. More recent models have leveraged attention mechanisms and Transformer architectures for improved performance \cite{bukhsh2021processtransformer}, explored autoencoders~\cite{ni2022predicting}, adversarial frameworks \cite{taymouri2020predictive,deleoni2024fairness}, LLMs~\cite{DBLP:conf/bpm/PadellaFNL25} and online continual next activity prediction under concept drift~\cite{vanstraten2026efficient}, while PGTNet~\cite{vanderaa2024pgtnet} represents the state of the art for remaining time prediction. As this is a regression task rather than a classification task, we do not include PGTNet as a baseline in this work. Despite these advances, a general issue with deep learning models is that they require large datasets to generalize well (cf. Section~\ref{sec:introduction}). To ensure stable training, it is recommended to have at least one unique training sample for each parameter being estimated \cite{goodfellow2016deep}. As deep learning models estimate thousands or even millions of parameters during training and event logs generally contain only a few thousand traces, often with redundant samples, there is a need for an artificial increase of training data. 

\subsection{Summary} \label{subsec:summary}
Table~\ref{tab:methods_table} summarizes the related work across PPM, NLP and CV. As highlighted by the clear gap (green square) in the literature, while advanced contrastive and Siamese learning paradigms and context-aware data augmentations are standard in NLP and CV, their intersection remains underexplored in PPM. Existing PPM approaches either rely on random, semantically invalid augmentations or computationally expensive LLMs that suffer from hallucination. Crucially, to the best of our knowledge, no method leverages the potential of Siamese learning to overcome labeled data scarcity. This fundamental research gap motivates our proposed framework, SiamSA-PPM, which introduces statistically grounded transformations specifically designed to enable robust, process-aware Siamese representation learning for event logs.

\begin{landscape}
\begin{table}[p]
  \centering
  \footnotesize 
  \setlength{\tabcolsep}{2pt} 
  \renewcommand{\arraystretch}{1.1} 
    \begin{tabular}{cll|cccc|cc|cc|c}
         \toprule
         \multicolumn{1}{c}{\rotatebox[origin=c]{70}{\textbf{Domain}}} & 
         \multicolumn{1}{c}{\rotatebox[origin=c]{70}{\begin{tabular}{@{}c@{}}\textbf{Method} \\ \textbf{(Authors)}\end{tabular}}} & 
         \multicolumn{1}{c|}{\rotatebox[origin=c]{70}{\textbf{Technique}}} & 
         \multicolumn{1}{c}{\rotatebox[origin=c]{70}{\scriptsize \begin{tabular}{@{}c@{}}\textbf{Process} \\ \textbf{Prediction}\end{tabular}}} & 
         \multicolumn{1}{c}{\rotatebox[origin=c]{70}{\scriptsize \begin{tabular}{@{}c@{}}\textbf{Image} \\ \textbf{Classification}\end{tabular}}} & 
         \multicolumn{1}{c}{\rotatebox[origin=c]{70}{\scriptsize \begin{tabular}{@{}c@{}}\textbf{Text} \\ \textbf{Classification}\end{tabular}}} & 
         \multicolumn{1}{c|}{\rotatebox[origin=c]{70}{\scriptsize \begin{tabular}{@{}c@{}}\textbf{Representation} \\ \textbf{Learning}\end{tabular}}} & 
         \multicolumn{1}{c}{\rotatebox[origin=c]{70}{\scriptsize \begin{tabular}{@{}c@{}}\textbf{Contrastive} \\ \textbf{Learning}\end{tabular}}} & 
         \multicolumn{1}{c|}{\rotatebox[origin=c]{70}{\scriptsize \begin{tabular}{@{}c@{}}\textbf{Siamese} \\ \textbf{Learning}\end{tabular}}} & 
         \multicolumn{1}{c}{\rotatebox[origin=c]{70}{\scriptsize \begin{tabular}{@{}c@{}}\textbf{Random} \\ \textbf{Augment.}\end{tabular}}} & 
         \multicolumn{1}{c}{\rotatebox[origin=c]{70}{\scriptsize \begin{tabular}{@{}c@{}}\textbf{Context-} \\ \textbf{aware}\end{tabular}}} & 
         \multicolumn{1}{c}{\rotatebox[origin=c]{70}{\scriptsize \begin{tabular}{@{}c@{}}\textbf{Hallucin-} \\ \textbf{ation}\end{tabular}}} \\ 
         \midrule
         
         \multirow{11}{*}{\rotatebox[origin=c]{90}{\textbf{PPM}}}
         & Tax et al. (2017)~\cite{tax2017predictive} & LSTM & \checkmark & & & & \cellcolor{green!15} & \cellcolor{green!15} & \cellcolor{green!15} & \cellcolor{green!15} & \cellcolor{green!15} \\ 
         & Di Mauro et al. (2019)~\cite{dimauro2019activity} & CNN & \checkmark & & & & \cellcolor{green!15} & \cellcolor{green!15} & \cellcolor{green!15} & \cellcolor{green!15} & \cellcolor{green!15} \\ 
         & Pasquadibisceglie et al. (2019)~\cite{pasquadibisceglie2019cnn} & CNN & \checkmark & & & & \cellcolor{green!15} & \cellcolor{green!15} & \cellcolor{green!15} & \cellcolor{green!15} & \cellcolor{green!15} \\ 
         & Bukhsh et al. (2021)~\cite{bukhsh2021processtransformer} & Transformer & \checkmark & & & & \cellcolor{green!15} & \cellcolor{green!15} & \cellcolor{green!15} & \cellcolor{green!15} & \cellcolor{green!15} \\ 
         & Ni et al. (2022)~\cite{ni2022predicting} & Auto-Encoder & \checkmark & & & & \cellcolor{green!15} & \cellcolor{green!15} & \cellcolor{green!15} & \cellcolor{green!15} & \cellcolor{green!15} \\ 
         & Taymouri et al. (2020)~\cite{taymouri2020predictive} & GAN & \checkmark & & & & \cellcolor{green!15} & \cellcolor{green!15} & \cellcolor{green!15} & \cellcolor{green!15} & \cellcolor{green!15} \\ 
         & De Leoni \& Padella (2024)~\cite{deleoni2024fairness} & GAN & \checkmark & & & & \cellcolor{green!15} & \cellcolor{green!15} & \cellcolor{green!15} & \cellcolor{green!15} & \cellcolor{green!15} \\ 
         & Padella et al. (2025)~\cite{DBLP:conf/bpm/PadellaFNL25} & LLM & \checkmark & & & & \cellcolor{green!15} & \cellcolor{green!15} & \cellcolor{green!15} & \cellcolor{green!15} & \cellcolor{green!15} \\ 
         & Amiri Elyasi et al. (2024)~\cite{vanderaa2024pgtnet} & PGTNet & \checkmark & & & & \cellcolor{green!15} & \cellcolor{green!15} & \cellcolor{green!15} & \cellcolor{green!15} & \cellcolor{green!15} \\ 
         & K\"appel \& Jablonski (2023)~\cite{kappel2023augmentation} & Random Augment. & \checkmark & & & & \cellcolor{green!15} & \cellcolor{green!15} & \cellcolor{green!15}\checkmark & \cellcolor{green!15} & \cellcolor{green!15} \\ 
         & K\"appel et al. (2026)~\cite{KAPPEL2026102717} & LLM Augment. & \checkmark & & & & \cellcolor{green!15} & \cellcolor{green!15} & \cellcolor{green!15}\checkmark & \cellcolor{green!15}\checkmark & \cellcolor{green!15}\checkmark \\ 
         \midrule
         
         \multirow{6}{*}{\rotatebox[origin=c]{90}{\textbf{NLP}}}
         & Wei \& Zou (2019)~\cite{wei2019eda} & EDA & & & \checkmark & & & & \checkmark & \checkmark & \\ 
         & Chou et al. (2020)~\cite{chou2020remix} & ReMix & & & \checkmark & & & & & \checkmark & \\ 
         & Lee et al. (2020)~\cite{lee2020smoothmix} & SmoothMix & & & \checkmark & & & & & \checkmark & \\ 
         & Ji et al. (2025)~\cite{Ji2025EMNLP} & Text-To-Query & & & \checkmark & & & & & \checkmark & \\ 
         & Wang et al. (2025)~\cite{Wang2025ACL} & DoAug & & & \checkmark & & & & & \checkmark & \\ 
         & Neuberger et al. (2026)~\cite{neuberger2026repeat} & Process IE Augment. & & & $\sim$ & & & & \checkmark & \checkmark & \\
         \midrule
         \multirow{9}{*}{\rotatebox[origin=c]{90}{\textbf{CV}}}
         & Chen et al. (2020)~\cite{chen2020simclr} & SimCLR & & \checkmark & & \checkmark & \checkmark & & \checkmark & & \\ 
         & Grill et al. (2020)~\cite{grill2020byol} & BYOL & & \checkmark & & \checkmark & & \checkmark & \checkmark & & \\ 
         & Chen et al. (2021)~\cite{chen2021simsiam} & SimSiam & & \checkmark & & \checkmark & & \checkmark & \checkmark & & \\ 
         & Mi \& Yang (2025)~\cite{Mi2025CVPR} & ADD & & \checkmark  & & & & & & \checkmark & \\ 
         & Yan et al. (2025)~\cite{Yan2025CVPR} & FedRDN & & \checkmark & & & & & & \checkmark & \\ 
         & Zhang et al. (2025)~\cite{Zhang2025CVPR} & GHEA & & \checkmark & & & & & & \checkmark & \\ 
         & Das et al. (2025)~\cite{Das2025CVPR} & COBRA & & \checkmark & & & & & & \checkmark & \\ 
         & Rahimi et al. (2025)~\cite{Rahimi2025NeurIPS} & AugGEN & & \checkmark & & & & & & \checkmark & \\ 
         & Kim et al. (2025)~\cite{Kim2025NeurIPS} & MRCA & & \checkmark & & & & & & \checkmark & \\ 
         \bottomrule
    \end{tabular}
    
    \vspace{0.2cm}
    \caption{Overview of related methods across Predictive Process Monitoring (PPM), Natural Language Processing (NLP), and Computer Vision (CV). The light green shaded block in the top right highlights the primary research gap addressed in this work: while representation learning (including contrastive and Siamese architectures) and advanced data augmentation techniques are highly prevalent in NLP and CV, they remain underexplored in PPM. Crucially, no existing PPM method integrates contrastive or Siamese learning with event log augmentation.}
    \label{tab:methods_table}
\end{table}
\end{landscape}
\section{Problem Definition}
\label{sec:problem_definition}

Let $L = \{ \sigma_1, \sigma_2, \dots, \sigma_N \}$ denote an event log composed of $N$ traces, where each trace $\sigma_i = \langle e_{i,1}, e_{i,2}, \dots, e_{i,T_i} \rangle$ is a sequence of $T_i$ events associated with a process instance. Each event $e_{i,t}$ carries attributes such as the activity label, timestamp and possibly additional case or event-level features. Each event $e_{i,t}$ is here intended as a tuple $(A_{i,t}, t_{i,t}, \vec{f}_{i,t})$, in which the activity $A_{i,t}\in\mathcal{A}$ for $\mathcal{A}$ the set of possible activities of the event log $L$,  $t_{i,t}\in\mathcal{T}$ the set of admitted timestamps and $\vec{f}\in\mathcal{F}_1\times\ldots\times\mathcal{F}_m$, where $\mathcal{F}_1,\ldots,\mathcal{F}_m$ are the sets containing the values of the different features. To lighten the notation, in this paper we are going to refer to $\langle A_1, \ldots, A_n \rangle$ as the sequence of events $\langle e_1, \ldots, e_n \rangle $ for which the corresponding activities are $A_1,\ldots, A_n$ and both the timestamps and the feature vectors are not known. In line with common practice in PPM, we treat each recorded event as the completion of an activity. Furthermore, given a trace $\sigma_i$, for $k\leq T_i$ we define $\sigma_i^{(k)} = \langle e_{i,1},e_{i,2},\ldots,e_{i,k} \rangle$ as the prefix that contains the first $k$ events of $\sigma_i$. The multiset comprising all such prefixes of all traces in an event log will be denoted by $\text{prefix}(L)$, i.e., $\text{prefix}(L) = \uplus_{\sigma_i \in L} \text{prefix}(\sigma_i)$.\footnote{The symbol $\uplus$ denotes the union of multisets.} Now, it is possible to define the prediction problems as follows:

\begin{enumerate}
    \item \textbf{Next Activity Prediction:} Given $\sigma^{(k)}_i$, predict $\hat{A}_{k+1,t}$ that should ideally match the ground truth activity label $A_{k+1,t}$.

    \item \textbf{Final Outcome Prediction:} Given $\sigma^{(k)}_i$, predict if there exists an event $e_{\hat{k}, i}\in\langle e_{k+1,i},\ldots,e_{n,i} \rangle$ such that $e_{\hat{k}, i}=(t, \hat{A}_{out}, \vec{f})$ for some $t$ and $\vec{f}$ where $\hat{A}_{out}$ should ideally match $A_{out}$ the ground truth final outcome label.
\end{enumerate}   


\section{Method}
\label{sec:method}

In this section, we present our methodology for event log augmentation and pre-training a robust encoder for the PPM tasks introduced in Section~\ref{sec:problem_definition}. In Section~\ref{subsec:novel_transformation_methods} we first introduce three statistically grounded trace transformation techniques based on activity sequences, while Sections~\ref{subsec:data_augmentation_strategy} and~\ref{subsec:siamese_pre_training} outline the different data augmentation pipelines and Siamese pre-training strategy, respectively. Finally, the fine-tuning procedure is introduced in Section~\ref{subsec:fine_tuning}.

\subsection{Novel Transformation Methods}\label{subsec:novel_transformation_methods}
\begin{figure}[h]
    \centering
    \includegraphics[width=0.9\linewidth]{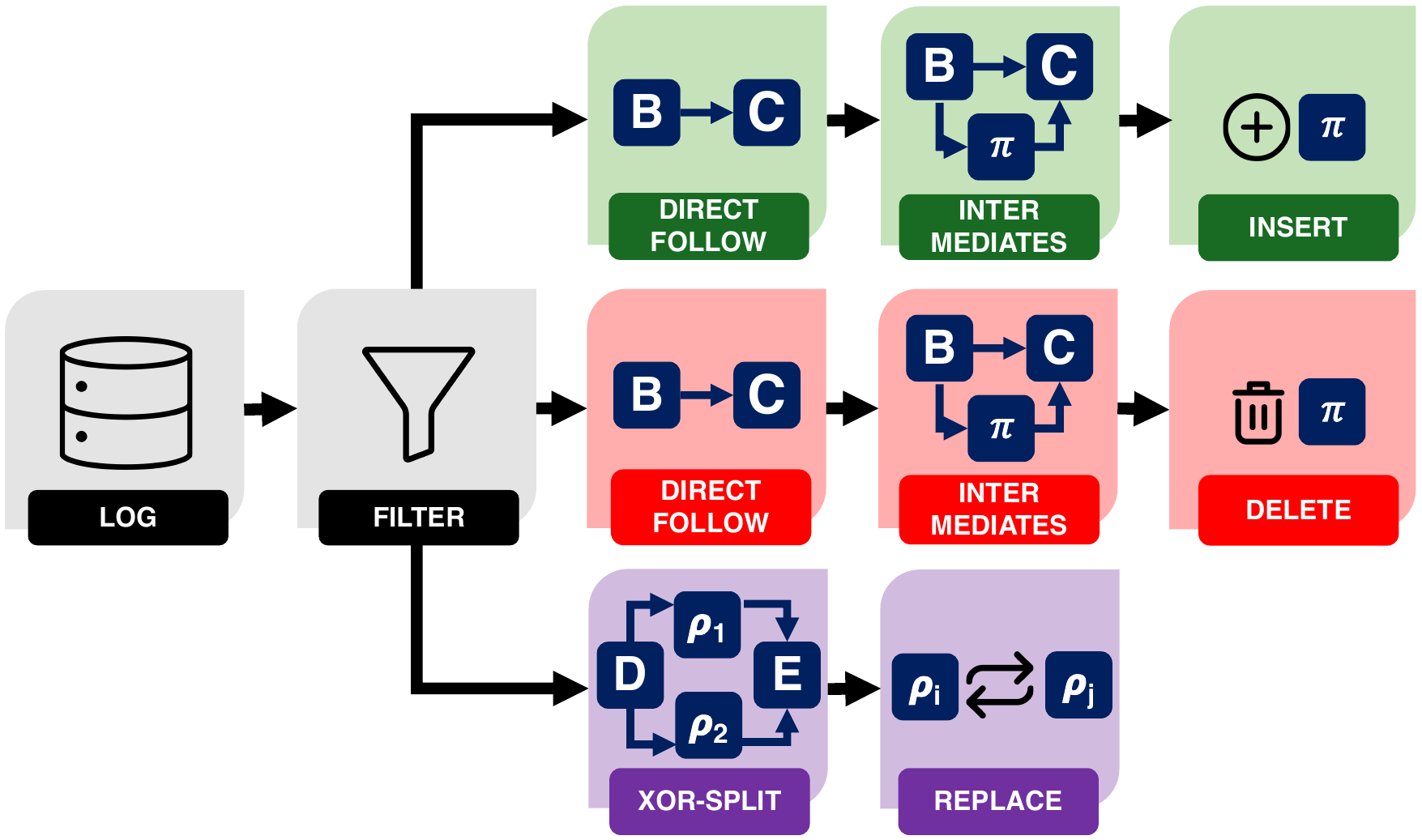}
    \caption{Overview of our three novel transformation methods: StatisticalInsertion (green), StatisticalDeletion (red) and StatisticalReplacement (purple).}
    \label{fig:novel_transformations}
\end{figure}

To keep the semantics of process traces and effectiveness of data augmentation in PPM, we propose three statistically grounded transformation techniques, named StatisticalInsertion, StatisticalDeletion and StatisticalReplacement (cf. Fig.~\ref{fig:novel_transformations}). Unlike traditional random transformations, these methods leverage frequent activity patterns in the event log to generate realistic and process-compliant trace variants. The techniques are parameterized by thresholds $\alpha, \beta, \gamma, \delta \in [0,1]$ and a maximum intermediate length $\lambda_{\max}$, which control frequency constraints and structural complexity.

Both \textit{StatisticalInsertion} and \textit{StatisticalDeletion} operate on observed direct and intermediate activity sequences. The procedure comprises the following steps. \textbf{1) Activity filtering.} From the original training log $L$, we construct a filtered log $L' \subseteq L$ by retaining only traces composed of activities that occur in at least factor $\alpha$ of all cases. This is solely needed for preprocessing and depends on the associated prediction task. \textbf{2) Direct follower extraction.} From $L'$, we extract all frequent direct follower pairs $(B \rightarrow C)$ where activity $C$ directly follows $B$ in at least factor $\beta$ of all observed transitions. These pairs represent commonly co-occurring dependencies. \textbf{3) Intermediate sequence mining.} For each direct pair $(B \rightarrow C)$, we identify frequent intermediate subsequences $\pi = \langle A_1, \dots, A_k \rangle$ such that $B \rightarrow \pi \rightarrow C$ occurs in at least factor $\gamma$ of traces in $L'$, with $1 \leq k \leq \lambda_{\max}$. \textbf{4) Consistency check.} Only those intermediate patterns $\pi$ are retained for which both $(B \rightarrow C)$ and $B \rightarrow \pi \rightarrow C$ satisfy the required frequency constraints. This ensures semantic consistency. \textbf{5) Transformation.} In the transformation phase, the selected patterns are applied to traces in the original log $L$. For insertion, if a trace $\sigma \in L$ contains a frequent direct follower $(B \rightarrow C)$, it may be replaced with the extended sequence $B \rightarrow \pi \rightarrow C$ for some eligible intermediate sequence $\pi$. Conversely, for deletion, if $\sigma$ contains a subsequence of the form $B \rightarrow \pi \rightarrow C$, this may be shortened to the corresponding direct follower $(B \rightarrow C)$. In both cases, the resulting traces preserve the behavioral structure derived from statistically grounded patterns observed in the training log.\\

{\textit{StatisticalReplacement} generalizes process behavior by identifying interchangeable subsequences between shared start and end points, targeting XOR-like structures. \textbf{1) Activity filtering.} The log is filtered as in the \textit{StatisticalInsertion} and \textit{StatisticalDeletion}. \textbf{2) Extract XOR-structures.} Next, we extract frequent patterns of the form $D \rightarrow \rho_i \rightarrow E$, where $\rho_i = \langle A_1, \dots, A_k \rangle$ is an intermediate subsequence of at most $\lambda_{\max}$ activities and each such subsequence must appear in equally or more than factor $\delta$ of traces. \textbf{3) Construct replacement set.} For each fixed pair of start and end points $(D, E)$, the various intermediate alternatives are collected into a replacement set $\mathcal{R}_{(D,E)} = \{ \rho_1, \rho_2, \dots, \rho_n \}$. \textbf{4) Transformation.} During augmentation, if a trace $\sigma \in L$ contains a segment matching $D \rightarrow \rho_i \rightarrow E$, the intermediate part $\rho_i$ may be substituted with a different alternative $\rho_j \in \mathcal{R}_{(D,E)}$, provided $\rho_j \neq \rho_i$.

For instance, in our running example, \textit{StatisticalInsertion} might add a frequent step $CheckDocs$ between $Order$ and $Pack$, while \textit{StatisticalReplacement} could substitute one observed path $Pack \rightarrow Ship \rightarrow Deliver$ with a special or quick delivery path $Pack \rightarrow ExpressShip \rightarrow Deliver$.

\subsection{Data Augmentation Strategy}\label{subsec:data_augmentation_strategy}
To generate diverse yet semantically coherent process trace variants, we apply two successive augmentations to each training prefix using a pool of transformation functions $\mathcal{Z}$. Our pipeline consists of two core stages: i) \textbf{applicability filtering} and ii) a \textbf{pairwise augmentation check}. Given a prefix $x$ of a trace, we first determine which augmentors are applicable by evaluating each method's structural constraints. Our main augmentors (\textit{StatisticalInsertion}, \textit{StatisticalDeletion} and \textit{StatisticalReplacement}) are applied when the sequence matches known behavioral patterns mined from the training log (cf. Section~\ref{subsec:novel_transformation_methods}). If none are applicable, we fall back to simpler, structure-agnostic augmentors named \textit{RandomInsertion}, \textit{RandomDeletion} and \textit{RandomReplacement}. For each input prefix $x$, we record all augmentors that are valid candidates for transformation. Next, we generate two augmented views \( v = t(x), v' = t'(x) \), where \( t, t' \in \mathcal{Z} \) are transformation functions sampled from the augmentation pool \( \mathcal{Z} \), such that \( v \neq v' \).
 We first sample an applicable augmentor uniformly and apply it to produce $v$. Then, we attempt up to 30 trials to generate a second distinct augmentation $v'$, avoiding trivial duplicates. Each augmented pair is padded using left-padding to a fixed length determined by the longest transformed sequence in the batch. This ensures input consistency for self-supervised learning. This strategy ensures each trace contributes semantically meaningful variation to the training set, leveraging both structural knowledge (from frequent patterns) and generalization capability (via fallback augmentors).

 For instance, in our running example, the prefix $Order \rightarrow Pack$ allows only insertion as valid transformation. Among the eligible candidates (derived from \textit{StatisticalInsertion}), both $CheckDocs$ and $ValidatePayment$ can be inserted, yielding $Order \rightarrow CheckDocs \rightarrow Pack$ and $Order \rightarrow ValidatePayment \rightarrow Pack$ as the two augmented prefixes. This illustrates how the strategy selects multiple valid options to generate distinct yet semantically consistent training views.

\subsection{Siamese Pre-Training}\label{subsec:siamese_pre_training}

\begin{figure}[h]
    \centering
    \includegraphics[width=0.99\linewidth]{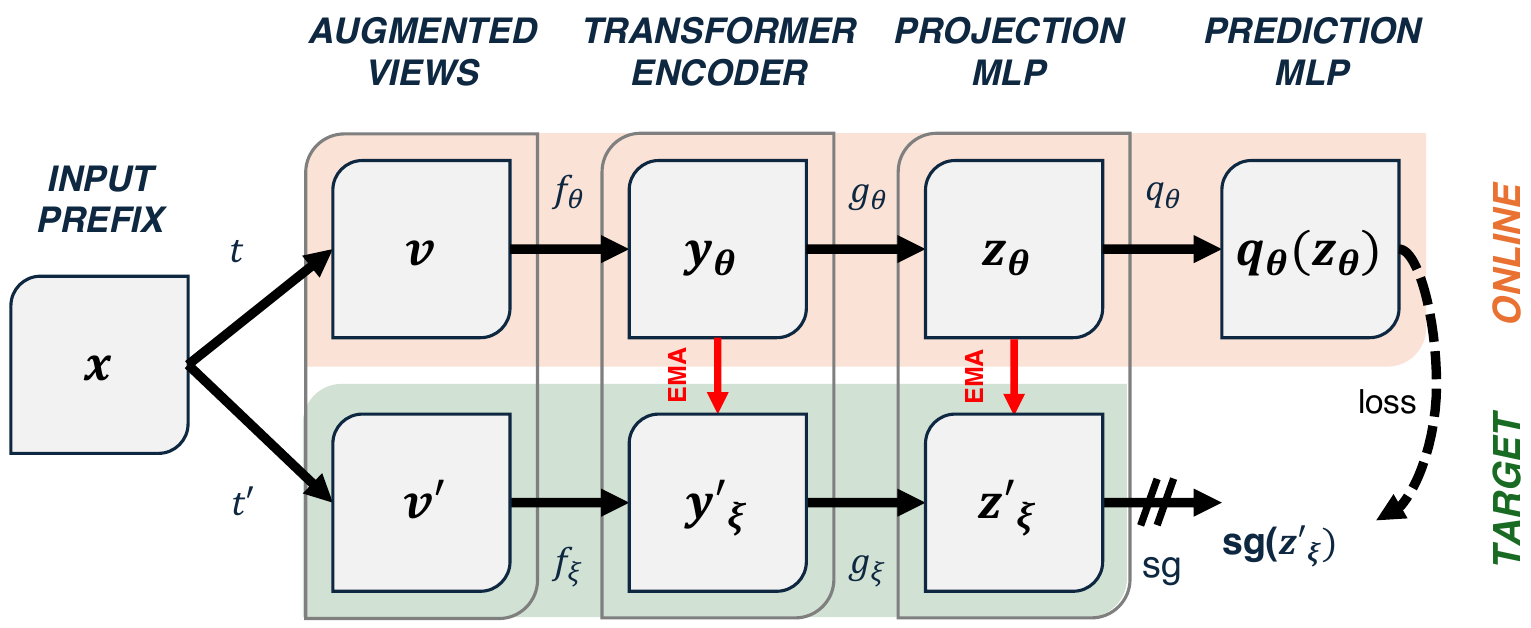}
    \caption{Overview of the pre-training architecture for SiamSA-PPM. Given an input prefix \( x \), two augmented views \( v \) and \( v' \) are generated using transformation functions $t$ and $t'$ chosen following our augmentation strategy and passed through a shared Transformer encoder and projection Multi-Layer Perceptron (MLP). The online network produces a prediction \( q_\theta(z_\theta) \), while the target network (updated via EMA) produces \( z'_\xi \). A loss is computed between the prediction and the stop-gradient of the target to enable representation alignment.}
    \label{fig:byol_architecture}
\end{figure}

To enable effective representation learning for process prefixes without requiring supervision, we adopt a Siamese learning framework inspired by BYOL~\cite{grill2020byol}. This self-supervised strategy facilitates the extraction of high-quality embeddings that capture structural aspects of process behavior. As illustrated in Fig.~\ref{fig:byol_architecture}, the training architecture consists of two neural branches: the \textit{online} network (top) and the \textit{target} network (bottom). Both receive different augmented views $v = t(x)$ and $v' = t'(x)$ of the same input prefix $x$, generated according to our data augmentation strategy (cf. Section \ref{subsec:data_augmentation_strategy}). These views are passed through a shared Transformer encoder backbone:

\begin{equation}
    y_\theta = f_\theta(v), \quad y'_\xi = f_\xi(v')
\end{equation}

Each encoded representation is then mapped to a latent space via a projection MLP: 

\begin{equation}
    z_\theta = g_\theta(y_\theta), \quad z'_\xi = g_\xi(y'_\xi)
\end{equation}

The online branch continues with a prediction MLP $q_\theta$, which is a small feedforward network that maps the projected representation $z_\theta$ to a predicted embedding $\hat{z}_\theta = q_\theta(z_\theta)$. This predictor is essential to avoid representational collapse: it allows the online network to learn a transformation of its own embeddings that can match the slowly evolving target embeddings. The target branch output $z'_\xi$ is detached from the computational graph using the stop-gradient operator $\operatorname{sg}(\cdot)$ and the loss is computed as the similarity between the predicted and target projections:

\begin{equation}
    \mathcal{L}_{\theta,\xi} = 2 - 2 \cdot \frac{q_\theta(z_\theta) \cdot \operatorname{sg}(z'_\xi)}{\|q_\theta(z_\theta)\|_2 \cdot \|\operatorname{sg}(z'_\xi)\|_2}
\label{loss_equation}
\end{equation}

We symmetrize the loss $\mathcal{L}_{\theta,\xi}$ in Eq.~\ref{loss_equation} by separately feeding $v'$ to the online network and $v$ to the target network to compute $\tilde{\mathcal{L}}_{\theta,\xi}$. The final training objective is the symmetric loss:

\begin{equation}
\mathcal{L}_{\theta,\xi} = \mathcal{L}_{\theta,\xi} + \tilde{\mathcal{L}}_{\theta,\xi}
\end{equation}

where each term compares the online network's prediction for one view to the target network's projection of the other. At each training step, we perform a stochastic optimization step to minimize $\mathcal{L}_{\theta,\xi}$ with respect to $\theta$ only, but \emph{not} $\xi$.

To avoid collapse (trivial solutions), \cite{chen2021simsiam} claims that only the stop-gradient operation is enough. However, our tests suggest otherwise, as the model collapses without the use of a momentum encoder (introduced in \cite{grill2020byol}). Hence, the target network parameters $\xi$ are not updated by gradient descent but rather using an exponential moving average (EMA) of the online network parameters: 

\begin{equation}
    \xi \leftarrow \tau \cdot \xi + (1 - \tau) \cdot \theta
\end{equation}

where $\tau \in [0, 1)$ is the momentum coefficient. This setup ensures that the model learns to align the online representation with a slowly evolving target representation, allowing robust and stable learning even in the absence of negative samples.

\subsection{Fine-Tuning}\label{subsec:fine_tuning}
Following pre-training, we discard the projection and prediction MLPs, but retain the encoder $f_\theta$ for downstream prediction tasks using labeled data, i.e. next activity prediction and final outcome prediction. We append a softmax classification layer to the encoder output to predict the next event or final outcome. The entire model, including the encoder, is fine-tuned end-to-end using cross-entropy loss. This allows the model to adapt its representations to the specific tasks while benefiting from the robust structure learned during pre-training. After fine-tuning, we evaluate the trained model.
\section{Experimental Setup}
\label{sec:experiments}
In this section, we present the datasets, evaluation metrics, competitors and implementation details required to assess the validity of the proposed approach.

\subsection{Datasets} \label{subsec:datasets}
We evaluate our approach on eight publicly available real-life case studies widely used in the process mining community \footnote{The majority of datasets are available at \protect\url{https://data.4tu.nl}. The BAC dataset is available at \protect\url{https://github.com/IBM/processmining}}. A core motivation of this work is to address training challenges in scenarios with limited data or low variability. Consequently, we categorize our dataset selection into two distinct groups: a \textit{target group} of smaller or structured logs where data augmentation is expected to be most beneficial, and a set of \textit{highly-variable} datasets used to validate competitiveness in data-rich environments. Table~\ref{tab:dataset_stats} summarizes the statistics for all selected logs, including the number of events, cases, activities, average trace length, and count of distinct trace variants (the main characteristic to divide the groups).

The \textit{target group} consists of datasets characterized by limited size or lower behavioral variability, representing the primary use case for our SiamSA-PPM framework. This category includes \textit{Helpdesk}, a log originating from the customer support system of an Italian software company, and a subset of the \textit{BPIC 2013} dataset from Volvo IT’s service management system: \textit{BPIC 2013-c} (closed problems). Additionally, we include \textit{Sepsis}, a healthcare log recording the trajectories of suspected sepsis patients and the \textit{Bank Account Closure (BAC)} dataset, a log from an Italian banking institution. Although this dataset is large in size, it has the least variability of all datasets that we evaluate on.

In contrast, to assess the scalability of our approach against state-of-the-art baselines, we evaluate performance on four benchmark datasets exhibiting high volume or variability. This group consists of \textit{BPIC 2012} and (the first 200,000 events of) \textit{BPIC 2017}, similar financial logs of personal loan and overdraft applications, noted for their complexity, as they capture} three intertwined subprocesses within a single event stream. To test performance on highly variable processes, we employ the first subset of \textit{BPIC 2015}, which covers building permit applications across five Dutch municipalities and contains numerous activity classes. Finally, we include the \textit{BPIC 2013-i} (incidents) dataset, a different subset of Volvo IT's service management system.


\begin{table}[t]
\centering
\caption{Statistics of the event logs used in the experimental evaluation, divided into the target group and highly variable datasets. The split is determined based on the number of variants.}
\label{tab:datasets}
\small
\setlength{\tabcolsep}{5pt}
\begin{tabular}{@{}cl rrrrr@{}}
\toprule
 & \textbf{Dataset} & \textbf{\#Events} & \textbf{\#Cases} & \textbf{\#Act.} & \textbf{Avg. Len.} & \textbf{\#Variants} \\
\midrule
\multirow{4}{*}{\rotatebox[origin=c]{90}{\textbf{Target}}}
 & \textit{Helpdesk} & 21,348  & 4,580  & 14  & 4.66  & 226   \\
 & \textit{BPIC13-c} & 6,660   & 1,487  & 7   & 4.48  & 327   \\
 & \textit{Sepsis}   & 15,214  & 1,050  & 16  & 14.49 & 846   \\
 & \textit{BAC}      & 212,721 & 32,429 & 15  & 6.56  & 189   \\
\midrule
\multirow{4}{*}{\rotatebox[origin=c]{90}{\textbf{Variable}}}
 & \textit{BPIC12}   & 262,200 & 13,087 & 36  & 20.04 & 4,366 \\
 & \textit{BPIC17}   & 200,033 & 5,168  & 25  & 38.71 & 3,248 \\
 & \textit{BPIC15-1} & 52,217  & 1,199  & 398 & 43.55 & 1,170 \\
 & \textit{BPIC13-i} & 65,533  & 7,554  & 13  & 8.68  & 2,278 \\
\bottomrule
\end{tabular}
\end{table}

\subsection{Metrics} \label{subsec:metrics}
Model performance is evaluated using four key metrics. \textit{Running time} measures the time taken for inference by computing the difference between the start and end timestamps during inference. \textit{Average accuracy} assesses overall predictive performance across the dataset (Eq.~\ref{eq:accuracy}): 

\begin{equation}\label{eq:accuracy}
    \text{Accuracy} = \frac{1}{\mathcal{N}} \sum_{j=1}^{\mathcal{N}} 1\{\hat{y}_j = y_j\}
\end{equation}

where \(\mathcal{N}\) is the total number of labels, \(\hat{y}_j\) the predicted and \(y_j\) the actual label, defined following the prediction problems introduced in Section~\ref{sec:problem_definition}. To assess the variability in augmented event logs, we adopt two entropy-based metrics introduced by \cite{back2019entropy}: \textit{trace entropy} and \textit{prefix entropy}.  Given a multiset $X$ of elements, where each element $x \in X$ is associated with a relative frequency $p(x)$, trace entropy is defined as Eq.~\ref{eq:trace_entropy}:

\begin{equation}\label{eq:trace_entropy}
    H_{trace} = - \sum_{\sigma \in L} p(\sigma) \log p(\sigma)
\end{equation}    

and captures the information contained in the event log in terms of trace variability and distribution. In a similar manner, prefix entropy considers the diversity of partial traces (prefixes), offering a finer-grained view of process variability by accounting for the workflow order of traces. It is computed as Eq.~\ref{eq:prefix_entropy}:

\begin{equation}\label{eq:prefix_entropy}
    H_{prefix} = - \sum_{\sigma' \in prefix(L)} p(\sigma') \log p(\sigma')
\end{equation}    

For next activity prediction, we report the average accuracy across the dataset. For final outcome prediction, we use binary classification targets and evaluate accuracy based on whether the predicted final outcome matches the actual label.

\subsection{Baselines} \label{subsec:baselines}
We compare our approach with four state-of-the-art methods for PPM. \textit{\textbf{Tax et al.}} \cite{tax2017predictive} use an LSTM-based model where each event is encoded as a feature vector including activity, time since the last event, time of day and weekday. The final hidden state of the LSTM summarizes the prefix and is passed to a dense layer to predict the next activity via softmax. \textit{\textbf{Di Mauro et al.}} \cite{dimauro2019activity} propose an inception-based 1D CNN that encodes activity and temporal features into a sequence. Inception modules apply multiple convolutions and pooling in parallel to capture varying local patterns, followed by global max pooling and classification. \textit{\textbf{Pasquadibisceglie et al.}} \cite{pasquadibisceglie2019cnn} transform prefixes into 2D matrices representing control-flow and time-performance data, processed by a standard CNN. This spatial encoding captures temporal patterns without recurrent structures, enabling efficient classification of future activities. \textit{\textbf{Bukhsh et al.}} \cite{bukhsh2021processtransformer} replace recurrence with a transformer encoder using self-attention to model dependencies across the trace. Learnable embeddings with positional encodings are processed through multi-head attention, followed by global pooling and dense layers for prediction.

\subsection{Implementation Details and Hyperparameters} \label{subsec:implementation_details}
The complete implementation of our approach, including data preprocessing pipelines, augmentation modules and PyTorch model architectures is publicly available on GitHub~\footnote{https://github.com/SvStraten/SiamSA-PPM-IJCIS}. All experiments are conducted in Python 3.12.7 on a MacBook Pro 2021 with Apple M1 processor (10-core CPU, 16-core GPU) and 16 GB Unified RAM, with 5 repetitions per experiment to address stochasticity of the generation process. We explicitly disabled parallel execution across experiments to accurately measure computational overhead and ensure consistent hardware resource allocation. To simulate a realistic deployment scenario and prevent future data leakage, we applied a strict temporal train-validation-test split of 65\%-15\%-20\% to all datasets.

Given the sensitivity of Siamese learning to the quality of generated views, we conducted an extensive hyperparameter grid search for each dataset. The grid search was evaluated based on the lowest validation loss during pre-training. We jointly optimized the hyperparameters governing both the statistical augmentation bounds and the neural network's training dynamics. Specifically, we evaluated the frequency thresholds ($\alpha$, $\beta$, $\gamma$, $\delta$) across the discrete search space of $\{10^{-3}, 10^{-4}\}$, and tested maximum path length constraints ($\lambda_{\text{max}}$) of 3 and 4. For the training dynamics, we explored batch sizes of $\{256, 512\}$ and base learning rates of $\{0.01, 0.05\}$. The optimal configurations for each event log are summarized in Table \ref{tab:hyperparameters}. Similarly, the augmentation factor used in the experiments of Section~\ref{subsec:augmentation_performance} was selected from $\{1.2, 1.5, 2.0\}$ based on validation accuracy. Interestingly, while the neural network hyperparameters converged to a consistent optimum across all logs ($bs = 256$, $\eta = 0.05$), the optimal statistical thresholds varied depending on the structural complexity and variability of the specific dataset. To ensure a fair and rigorous evaluation, we performed a comparable hyperparameter tuning process for all baseline methods. Their final configurations were similarly selected based on optimal validation performance to maximize their respective predictive capabilities.

\begin{table}[h]
    \centering
    \caption{Optimal hyperparameters selected for each event log after grid search on the validation set.}
    \label{tab:hyperparameters}
    \renewcommand{\arraystretch}{1.15}
    \setlength{\tabcolsep}{3pt}
    \resizebox{\textwidth}{!}{%
    \small
    \begin{tabular}{ll ccccccc cc cc cc cc}
    \toprule
    & & \multicolumn{7}{c}{\textbf{SiamSA-PPM (ours)}} & \multicolumn{2}{c}{\textbf{Tax~\cite{tax2017predictive}}} & \multicolumn{2}{c}{\textbf{Di Mauro~\cite{dimauro2019activity}}} & \multicolumn{2}{c}{\textbf{Pasquad.~\cite{pasquadibisceglie2019cnn}}} & \multicolumn{2}{c}{\textbf{Bukhsh~\cite{bukhsh2021processtransformer}}} \\
    \cmidrule(lr){3-9} \cmidrule(lr){10-11} \cmidrule(lr){12-13} \cmidrule(lr){14-15} \cmidrule(lr){16-17}
    & \textbf{Dataset} & $\boldsymbol{\alpha}$ & $\boldsymbol{\beta}$ & $\boldsymbol{\gamma}$ & $\boldsymbol{\delta}$ & $\boldsymbol{\lambda}_{\textbf{max}}$ & \textbf{bs} & $\boldsymbol{\eta}$ & $\boldsymbol{units}$ & $\boldsymbol{\eta}$ & $\boldsymbol{filters}$ & $\boldsymbol{\eta}$ & $\boldsymbol{filters}$ & $\boldsymbol{\eta}$ & $\boldsymbol{emb}$ & $\boldsymbol{\eta}$ \\
    \midrule
    \multirow{4}{*}{\rotatebox[origin=c]{90}{\textbf{Target}}} 
    & \textit{Helpdesk} & $10^{-3}$ & $10^{-3}$ & $10^{-3}$ & $10^{-4}$ & 4 & 256 & 0.05 & 150 & 0.0005 & 16 & 0.0005 & 16 & 0.001 & 32 & 0.0005 \\ 
    & \textit{BPIC13-c} & $10^{-3}$ & $10^{-4}$ & $10^{-4}$ & $10^{-3}$ & 4 & 256 & 0.05 & 100 & 0.0005 & 16 & 0.001 & 16 & 0.001 & 32 & 0.0005 \\ 
    & \textit{Sepsis}   & $10^{-3}$ & $10^{-4}$ & $10^{-4}$ & $10^{-3}$ & 3 & 256 & 0.05 & 50 & 0.001 & 64 & 0.001 & 32 & 0.001 & 128 & 0.001 \\ 
    & \textit{BAC}      & $10^{-4}$ & $10^{-4}$ & $10^{-4}$ & $10^{-4}$ & 3 & 256 & 0.05 & 50 & 0.001 & 32 & 0.0005 & 32 & 0.0005 & 128 & 0.001 \\
    \midrule
    \multirow{4}{*}{\rotatebox[origin=c]{90}{\textbf{Variable}}} 
    & \textit{BPIC12}   & $10^{-4}$ & $10^{-4}$ & $10^{-3}$ & $10^{-4}$ & 4 & 256 & 0.05 & 100 & 0.001 & 64 & 0.001 & 32 & 0.0001 & 64 & 0.001 \\ 
    & \textit{BPIC17}   & $10^{-3}$ & $10^{-3}$ & $10^{-4}$ & $10^{-3}$ & 4 & 256 & 0.05 & 100 & 0.0005 & 64 & 0.0005 & 64 & 0.001 & 128 & 0.001 \\ 
    & \textit{BPIC15-1} & $10^{-3}$ & $10^{-4}$ & $10^{-4}$ & $10^{-4}$ & 3 & 256 & 0.05 & 150 & 0.0005 & 64 & 0.0005 & 16 & 0.0005 & 128 & 0.001 \\ 
    & \textit{BPIC13-i} & $10^{-4}$ & $10^{-3}$ & $10^{-3}$ & $10^{-4}$ & 4 & 256 & 0.05 & 150 & 0.001 & 32 & 0.0005 & 64 & 0.001 & 64 & 0.001 \\
    \bottomrule
    \end{tabular}%
    }
\end{table}
\section{Evaluation Results}
\label{sec:results}

\subsection{Entropy} \label{subsec:entropy}
To quantitatively assess the impact of our proposed data augmentation strategy (cf. Section~\ref{subsec:data_augmentation_strategy}) on the variability of the evaluated event logs, we analyze two entropy-based metrics: trace entropy and prefix entropy (cf. Section~\ref{subsec:metrics}). 

\begin{figure}[h]
    \centering
    \includegraphics[width=1\linewidth]{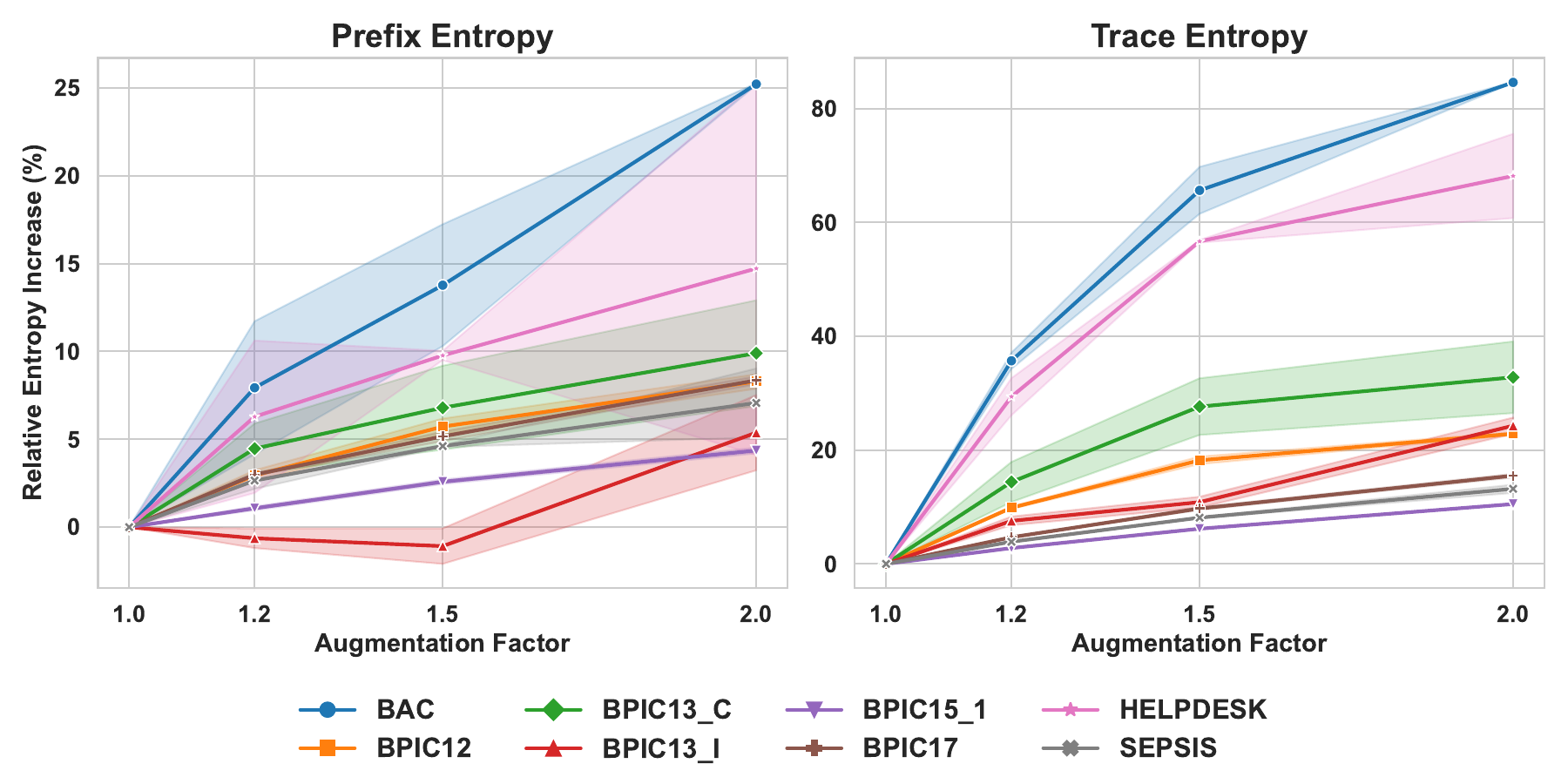}
    \vspace{-1em}
    \caption{Increase in prefix and trace entropy as a function of augmentation factor, showing how data augmentation enhances variability across different event logs.}
    \label{fig:entropy}
\end{figure}

Fig.~\ref{fig:entropy} illustrates the relative increase in both entropy metrics as a function of the augmentation factor, scaling from 1.0 (the raw, original training log), via 1.2 and 1.5 up to 2.0 (effectively doubling the size of the training log). The results demonstrate a consistent and positive correlation between the augmentation factor and the overall log variability across the majority of cases.

The most substantial entropy gains occur in our target group, particularly the \textit{BAC} and \textit{Helpdesk} logs. At an augmentation factor of 2.0, \textit{BAC}'s trace entropy surges by over 80\%, with prefix entropy rising around 25\%. This increase stems from \textit{BAC}'s inherent low variability, since it contains only 189 distinct trace variants among 32,429 cases, making it the least variable dataset evaluated (cf. Table~\ref{tab:dataset_stats}). Consequently, our statistically grounded transformations successfully inject new, semantically valid patterns into the augmented log. A similar trend applies to the \textit{Helpdesk} log, which natively has just 226 variants. Conversely, entropy increases are more moderate for large-scale logs that already exhibit high complexity. For example, \textit{BPIC15-1} is highly variable, naturally containing 1,170 variants across only 1,199 cases. In such already-complex logs, the relative impact of augmentation is inherently lower, as can be seen in Fig.~\ref{fig:entropy}. One exception is the \textit{BPIC13-i} dataset, whose prefix entropy slightly decreases at factors 1.2 and 1.5 before increasing at 2.0. Since prefix entropy is frequency weighted (cf. Eq.~\ref{eq:prefix_entropy}), augmented prefixes that duplicate frequent ones concentrate the distribution and can lower entropy. \textit{BPIC13-i} is likely prone to this, as it combines few activities (13) and short traces (8.68 on average), with many variants (2,278). At factor 2.0, enough new prefixes are introduced to outweigh this effect.

These findings confirm our core hypothesis: semantic-preserving augmentation measurably enhances process variability. By augmenting based on existing process behavior, our method provides a richer foundation for representation learning. The critical question, however, is whether this variability enrichment actually translates into tangible performance improvements when applied to standard predictive baselines.

\subsection{Effect of Augmentation on Performance} \label{subsec:augmentation_performance}
To isolate and evaluate the direct impact of our data augmentation strategy on predictive performance, we assess the average accuracy of the four state-of-the-art baselines across all eight datasets. Each baseline model is trained on three distinct configurations, namely the original training log, a training log augmented using the random transformation methods from~\cite{kappel2023augmentation} and a training log augmented with our proposed statistically grounded strategy (cf. Section~\ref{subsec:data_augmentation_strategy}). Crucially, all models are subsequently evaluated on the exact same, unseen test log. Table~\ref{tab:augmented_logs} reports the next activity prediction results using the best-performing augmentation factor (1.2, 1.5 or 2.0), selected for each log and baseline based on the accuracy on the validation set. The test set was not used for this selection.

On the target datasets, the effect of augmentation is mixed. Our strategy improves over the original training log in 9 of the 16 dataset and baseline combinations, compared to 7 of 16 for random augmentation. Gains are consistent on \textit{Sepsis}, where all four baselines improve, while half of the baselines improve on \textit{BPIC13-c} and \textit{BAC}. On \textit{Helpdesk}, however, three of the four baselines lose accuracy, and on \textit{BPIC13-c} the model of Pasquadibisceglie et al.~\cite{pasquadibisceglie2019cnn} drops by 6.62 percentage points. Random augmentation shows similar fluctuations, with accuracy drops for three of the four baselines on both \textit{Helpdesk} and \textit{BPIC13-c}. On the highly variable datasets, our strategy behaves more consistently. It improves over the original log in 12 of the 16 combinations, with one tie and only small decreases in the remaining cases. The largest gains are observed on \textit{BPIC15-1}, where all four baselines improve. We hypothesize that in low-variability logs, augmentation shifts the distribution over the few existing variants more strongly (cf. Section~\ref{subsec:entropy}), moving it away from the test distribution. Overall, our strategy maintains or improves performance on the highly variable logs, whereas its effect on the target datasets is less consistent.

\begin{table}[!htbp]
\centering
\caption{Accuracy of the baselines for next activity prediction using the original training log, the augmented training log following \protect\cite{kappel2023augmentation} and the augmented training log following our strategy (cf. Section \ref{subsec:data_augmentation_strategy}). \textbf{Bold} denotes the highest accuracy per dataset and baseline.} 
\label{tab:augmented_logs}
\renewcommand{\arraystretch}{1.2}
\resizebox{\textwidth}{!}{%
\small
\begin{tabular}{llcccc}
\toprule
\multirow{2}{*}{\textbf{Dataset}} & \multirow{2}{*}{\textbf{Log Type}} & \multicolumn{4}{c}{\textbf{Method}} \\
\cmidrule(lr){3-6}
 &  & \textit{\textbf{Tax \cite{tax2017predictive}}} 
  & \textit{\textbf{Di\,Mauro \cite{dimauro2019activity}}} 
  & \textit{\textbf{Pasquad.\ \cite{pasquadibisceglie2019cnn}}} 
  & \textbf{\textit{Bukhsh \cite{bukhsh2021processtransformer}}} \\
\midrule

\multirow{3}{*}{\textit{Helpdesk}} 
  & Original Log & $\bm{70.63} \pm \bm{0.44}$ & $\bm{78.85} \pm \bm{0.04}$ & $\bm{78.81} \pm \bm{0.17}$ & $78.45 \pm 0.25$ \\
  & + Random Augment.~\cite{kappel2023augmentation} & $69.21 \pm 0.68$ & $75.56 \pm 0.43$ & $78.64 \pm 0.32$ & $78.76 \pm 0.17$ \\
  & + Our Strategy & $68.79 \pm 0.78$ & $77.54 \pm 0.56$ & $78.71 \pm 0.12$ & $\bm{78.93} \pm \bm{0.10}$ \\
\midrule

\multirow{3}{*}{\textit{BPIC13-c}} 
  & Original Log & $\bm{55.18} \pm \bm{3.17}$ & $56.85 \pm 2.01$ & $\bm{52.31} \pm \bm{0.17}$ & $55.92 \pm 2.23$ \\
  & + Random Augment.~\cite{kappel2023augmentation} & $53.81 \pm 3.73$ & $57.89 \pm 1.97$ & $52.01 \pm 2.56$ & $54.97 \pm 2.69$ \\
  & + Our Strategy & $54.20 \pm 1.26$ & $\bm{58.19} \pm \bm{2.10}$ & $45.69 \pm 4.52$ & $\bm{56.72} \pm \bm{1.26}$ \\
\midrule

\multirow{3}{*}{\textit{Sepsis}} 
  & Original Log & $52.69 \pm 1.40$ & $59.44 \pm 1.93$ & $56.07 \pm 1.72$ & $59.16 \pm 2.18$ \\
  & + Random Augment.~\cite{kappel2023augmentation} & $52.89 \pm 1.81$ & $59.83 \pm 2.11$ & $55.93 \pm 2.97$ & $60.29 \pm 1.65$ \\
  & + Our Strategy & $\bm{52.91} \pm \bm{1.09}$ & $\bm{59.87} \pm \bm{1.27}$ & $\bm{56.39} \pm \bm{1.54}$ & $\bm{61.50} \pm \bm{1.12}$ \\
\midrule

\multirow{3}{*}{\textit{BAC}}
  & Original Log & $80.32 \pm 0.31$ & $95.12 \pm 0.20$ & $\bm{95.19} \pm \bm{0.22}$ & $\bm{95.47} \pm \bm{0.03}$ \\    
  & + Random Augment.~\cite{kappel2023augmentation} & $80.43 \pm 0.22$ & $95.15 \pm 0.09$ & $95.02 \pm 0.07$ & $95.34 \pm 0.24$ \\
  & + Our Strategy & $\bm{80.54} \pm \bm{0.22}$ & $\bm{95.21} \pm \bm{0.13}$ & $95.15 \pm 0.06$ & $95.18 \pm 0.31$ \\
\midrule[\heavyrulewidth] 

\multirow{3}{*}{\textit{BPIC12}}
  & Original Log & $\bm{70.85} \pm \bm{0.79}$ & $83.20 \pm 1.00$ & $83.14 \pm 0.27$ & $83.47 \pm 0.31$ \\
  & + Random Augment.~\cite{kappel2023augmentation} & $69.76 \pm 0.27$ & $\bm{83.46} \pm \bm{0.25}$ & $83.03 \pm 0.24$ & $\bm{83.75} \pm \bm{0.35}$ \\
  & + Our Strategy & $69.68 \pm 0.52$ & $83.23 \pm 0.29$ & $\bm{83.28} \pm \bm{0.06}$ & $83.65 \pm 0.54$ \\
\midrule

\multirow{3}{*}{\textit{BPIC17}}
  & Original Log & $75.25 \pm 0.11$ & $\bm{88.77} \pm \bm{0.13}$ & $88.48 \pm 0.23$ & $\bm{88.63} \pm \bm{0.12}$ \\
  & + Random Augment.~\cite{kappel2023augmentation} & $\bm{75.35} \pm \bm{0.08}$ & $88.41 \pm 0.12$ & $88.42 \pm 0.29$ & $88.40 \pm 0.06$ \\
  & + Our Strategy & $\bm{75.35} \pm \bm{0.12}$ & $88.72 \pm 0.03$ & $\bm{88.75} \pm \bm{0.09}$ & $\bm{88.63} \pm \bm{0.15}$ \\
\midrule

\multirow{3}{*}{\textit{BPIC15-1}} 
  & Original Log & $24.83 \pm 0.44$ & $47.15 \pm 0.55$ & $41.90 \pm 0.46$ & $38.54 \pm 1.53$ \\
  & + Random Augment.~\cite{kappel2023augmentation} & $22.12 \pm 3.44$ & $44.12 \pm 1.58$ & $42.31 \pm 1.21$ & $37.48 \pm 1.65$ \\
  & + Our Strategy & $\bm{29.34} \pm \bm{1.62}$ & $\bm{47.46} \pm \bm{0.36}$ & $\bm{42.78} \pm \bm{0.89}$ & $\bm{40.49} \pm \bm{0.55}$ \\
\midrule

\multirow{3}{*}{\textit{BPIC13-i}} 
  & Original Log & $55.02 \pm 0.51$ & $59.39 \pm 0.11$ & $\bm{59.10} \pm \bm{3.79}$ & $58.69 \pm 0.12$ \\
  & + Random Augment.~\cite{kappel2023augmentation} & $\bm{56.45} \pm \bm{1.01}$ & $59.98 \pm 1.67$ & $58.56 \pm 3.67$ & $57.96 \pm 1.32$ \\
  & + Our Strategy & $55.34 \pm 0.22$ & $\bm{61.24} \pm \bm{0.88}$ & $59.02 \pm 2.78$ & $\bm{58.98} \pm \bm{0.28}$ \\
\bottomrule 
\end{tabular}%
}
\end{table}

\subsection{Next Activity Prediction} \label{subsec:next_activity_prediction}
We evaluate SiamSA-PPM's predictive performance against the four state-of-the-art baselines. To ensure a fair comparison and isolate the architectural and pre-training benefits of our approach, all models are trained on the unaugmented training datasets and evaluated on the exact same, unseen test datasets. This allows us to assess baseline capabilities independent of explicit downstream augmentation.

As detailed in Table~\ref{tab:next_activity_target}, SiamSA-PPM excels on datasets characterized by labeled training data scarcity and low behavioral variability. Our method achieves the highest average accuracy on the \textit{BPIC13-c} (57.06\%) and \textit{Sepsis} (60.62\%) datasets, and performs on par with the best baseline on \textit{BAC} (95.49\% versus 95.47\% for Bukhsh et al.~\cite{bukhsh2021processtransformer}). On the \textit{Helpdesk} dataset, SiamSA-PPM remains highly competitive at 78.60\%, performing only marginally below the top baseline by Di Mauro et al. (78.85\%). This highlights the efficacy of our self-supervised framework. By extracting task-agnostic embeddings via a Siamese learning setup, our framework effectively mitigates the lack of natural variability. This allows the encoder to generalize better than standard supervised models, which are prone to overfitting when training data is limited.

\begin{table}[!htbp]
\centering
\caption{Accuracy on the target group datasets (mean $\pm$ std) for next activity prediction. \textbf{Bold} denotes the highest performance, \textit{italic} and \underline{underline} the second and third highest respectively.}
\label{tab:next_activity_target}
\footnotesize
\setlength{\tabcolsep}{4pt}
\begin{tabular}{@{}l ccccc@{}}
\toprule
\textbf{Dataset} & \textbf{Tax}~\cite{tax2017predictive} & \textbf{Di Mauro}~\cite{dimauro2019activity} & \textbf{Pasquad.}~\cite{pasquadibisceglie2019cnn} & \textbf{Bukhsh}~\cite{bukhsh2021processtransformer} & \textbf{SiamSA-PPM} \\
\midrule
\textit{Helpdesk} & $70.63 \pm 0.44$ & $\bm{78.85} \pm \bm{0.04}$ & $\mathit{78.81} \pm \mathit{0.17}$ & $78.45 \pm 0.25$ & $\underline{78.60} \pm \underline{0.18}$ \\
\textit{BPIC13-c} & $55.18 \pm 3.17$ & $\mathit{56.85} \pm \mathit{2.01}$ & $52.31 \pm 0.17$ & $\underline{55.92} \pm \underline{2.23}$ & $\bm{57.06} \pm \bm{0.24}$ \\
\textit{Sepsis}   & $52.69 \pm 1.40$ & $\mathit{59.44} \pm \mathit{1.93}$ & $56.07 \pm 1.72$ & $\underline{59.16} \pm \underline{2.18}$ & $\bm{60.62} \pm \bm{0.20}$ \\
\textit{BAC}      & $80.32 \pm 0.31$ & $95.12 \pm 0.20$ & $\underline{95.19} \pm \underline{0.22}$ & $\mathit{95.47} \pm \mathit{0.03}$ & $\bm{95.49} \pm \bm{0.01}$ \\
\bottomrule
\end{tabular}
\end{table}

\begin{table}[!htbp]
\centering
\caption{Accuracy on the highly variable datasets (mean $\pm$ std) for next activity prediction. \textbf{Bold} denotes the highest performance, \textit{italic} and \underline{underline} the second and third highest respectively.}
\label{tab:next_activity_large_scale}
\footnotesize
\setlength{\tabcolsep}{4pt}
\begin{tabular}{@{}l ccccc@{}}
\toprule
\textbf{Dataset} & \textbf{Tax}~\cite{tax2017predictive} & \textbf{Di Mauro}~\cite{dimauro2019activity} & \textbf{Pasquad.}~\cite{pasquadibisceglie2019cnn} & \textbf{Bukhsh}~\cite{bukhsh2021processtransformer} & \textbf{SiamSA-PPM} \\
\midrule
\textit{BPIC12}   & $70.85 \pm 0.79$ & $\mathit{83.20} \pm \mathit{1.00}$ & $\underline{83.14} \pm \underline{0.27}$ & $\bm{83.47} \pm \bm{0.31}$ & $79.48 \pm 0.05$ \\
\textit{BPIC17}   & $75.25 \pm 0.11$ & $\bm{88.77} \pm \bm{0.13}$ & $\underline{88.48} \pm \underline{0.23}$ & $\mathit{88.63} \pm \mathit{0.12}$ & $85.65 \pm 0.06$ \\
\textit{BPIC15-1} & $24.83 \pm 0.44$ & $\bm{47.15} \pm \bm{0.55}$ & $\mathit{41.90} \pm \mathit{0.46}$ & $\underline{38.54} \pm \underline{1.53}$ & $37.11 \pm 0.13$ \\
\textit{BPIC13-i} & $55.02 \pm 0.51$ & $\bm{59.39} \pm \bm{0.11}$ & $\underline{59.10} \pm \underline{3.79}$ & $58.69 \pm 0.12$ & $\mathit{59.27} \pm \mathit{0.08}$ \\
\bottomrule
\end{tabular}
\end{table}

Conversely, Table~\ref{tab:next_activity_large_scale} illustrates a different dynamic for highly-variable benchmark datasets. In these data-rich datasets, standard deep learning approaches have sufficient examples to optimize effectively, often establishing the state-of-the-art. For instance, Di Mauro et al. achieve the highest accuracy on \textit{BPIC13-i} (59.39\%), \textit{BPIC15-1} (47.15\%) and \textit{BPIC17} (88.77\%), while Bukhsh et al. have the highest accuracy on \textit{BPIC12} (83.47\%). SiamSA-PPM yields lower but still competitive results across these logs, which aligns with our expectations. The relative impact of self-supervised pre-training naturally diminishes when the training dataset already possesses high volume and structural complexity.

Overall, the evaluation indicates that SiamSA-PPM is effective for data-scarce real-world logs, while remaining competitive on most of the highly variable benchmark datasets.

\subsection{Final Outcome Prediction} \label{subsec:final_outcome_prediction}
In addition to next activity prediction, we evaluate the ability of our framework to forecast the final outcomes of ongoing process instances. Table~\ref{tab:final_outcome} summarizes the accuracy results of the binary classification targets across the evaluated methods. Unlike next activity prediction, final outcome prediction requires event logs that exhibit distinct and varying case resolutions. Several datasets, such as \textit{Helpdesk}, \textit{BPIC13-c} and \textit{BPIC13-i}, are excluded from this evaluation, as they predominantly contain a single, uniform resolution path. We focus our analysis on the \textit{BPIC12} and \textit{Sepsis} datasets, which provide clear, multi-faceted binary classification targets. As with the previous experiments, the baseline models are trained conventionally on the original training event logs, while SiamSA-PPM leverages representations via our augmented self-supervised pre-training.

For the \textit{BPIC12} dataset, the predictive targets correspond to the final status of the loan application requests, namely \texttt{Approved}, \texttt{Declined} or \texttt{Cancelled}. Even in this data-rich environment, SiamSA-PPM remains highly competitive. It achieves 80.04\% accuracy on the \texttt{Declined} target and 75.39\% on \texttt{Cancelled}, trailing the top-performing  Bukhsh et al. and Di Mauro et al. baselines by less than 1 percentage point in both cases. The advantages of our statistically grounded pre-training are more visible in the data-scarce \textit{Sepsis} dataset, where targets indicate the clinical justification for a patient's release (\texttt{Release-A} through \texttt{D}). Here, SiamSA-PPM achieves the highest predictive accuracy for both \texttt{Release-B} (94.21\%) and \texttt{Release-D} (96.02\%), while remaining highly competitive on \texttt{Release-C}.

\begin{table}[!htbp]
\centering
\caption{Accuracy (mean $\pm$ std) for final outcome prediction. \textbf{Bold} denotes the highest performance, \textit{italic} and \underline{underline} the second and third highest respectively.}
\label{tab:final_outcome}
\footnotesize
\setlength{\tabcolsep}{3.5pt}
\begin{tabular}{@{}cl ccccc@{}}
\toprule
 & \textbf{Target} & \textbf{Tax}~\cite{tax2017predictive} & \textbf{Di Mauro}~\cite{dimauro2019activity} & \textbf{Pasquad.}~\cite{pasquadibisceglie2019cnn} & \textbf{Bukhsh}~\cite{bukhsh2021processtransformer} & \textbf{SiamSA-PPM} \\
\midrule
\multirow{3}{*}{\rotatebox[origin=c]{90}{\textit{BPIC12}}}
 & \texttt{Approved}  & $65.63 \pm 3.40$ & $56.42 \pm 0.25$ & $\mathit{74.73} \pm \mathit{0.58}$ & $\bm{75.43} \pm \bm{0.22}$ & $\underline{73.58} \pm \underline{0.14}$ \\
 & \texttt{Declined}  & $75.31 \pm 1.50$ & $69.70 \pm 0.64$ & $\underline{79.50} \pm \underline{1.15}$ & $\bm{80.86} \pm \bm{0.04}$ & $\mathit{80.04} \pm \mathit{0.02}$ \\
 & \texttt{Cancelled} & $63.36 \pm 2.48$ & $\bm{75.76} \pm \bm{0.01}$ & $74.47 \pm 3.02$ & $\mathit{75.63} \pm \mathit{1.13}$ & $\underline{75.39} \pm \underline{0.06}$ \\
\midrule
\multirow{4}{*}{\rotatebox[origin=c]{90}{\textit{Sepsis}}}
 & \texttt{Release-A} & $\bm{84.15} \pm \bm{2.27}$ & $77.39 \pm 0.98$ & $\underline{81.95} \pm \underline{0.48}$ & $\mathit{83.42} \pm \mathit{0.83}$ & $80.71 \pm 0.11$ \\
 & \texttt{Release-B} & $92.47 \pm 1.75$ & $91.17 \pm 2.42$ & $\underline{93.28} \pm \underline{0.43}$ & $\mathit{94.06} \pm \mathit{0.65}$ & $\bm{94.21} \pm \bm{0.08}$ \\
 & \texttt{Release-C} & $90.58 \pm 0.11$ & $90.10 \pm 0.49$ & $\mathit{90.80} \pm \mathit{0.06}$ & $\bm{90.82} \pm \bm{0.00}$ & $\underline{90.59} \pm \underline{0.00}$ \\
 & \texttt{Release-D} & $94.94 \pm 1.27$ & $\mathit{95.81} \pm \mathit{0.58}$ & $\underline{95.64} \pm \underline{0.78}$ & $95.58 \pm 0.72$ & $\bm{96.02} \pm \bm{0.11}$ \\
\bottomrule
\end{tabular}
\end{table}

\subsection{Non-Deterministic Analysis}\label{subsec:non_deterministic}
Standard evaluation metrics, when applied over complete event sequences, can sometimes obscure the true impact and quality of data augmentation techniques due to the high frequency of deterministic, highly certain event sequences. To rigorously evaluate the semantic validity of our proposed augmentation strategy, we conducted an in-depth analysis focusing exclusively on non-deterministic decision points (XOR-splits). We evaluated the preservation of control-flow by comparing the augmented training logs to a held-out test set (20\%), demonstrating results across both flexible (\textit{BPIC12}) and highly structured (\textit{BAC}) datasets.

\begin{figure}[t]
    \centering
    \includegraphics[width=0.66\linewidth]{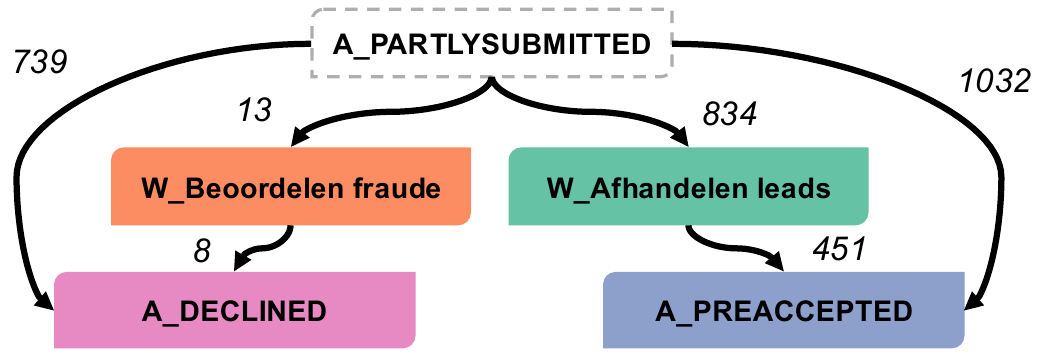}
    \caption{Next activity transitions following the \texttt{A\_PARTLYSUBMITTED} event in the \textit{BPIC12} dataset. The absolute transition counts of the test set (20\%) establish the ground truth.}
    \label{fig:test_bpic12}
\end{figure}

\begin{figure}[t]
   \vspace{-1em}
    \centering
    \includegraphics[width=1\linewidth]{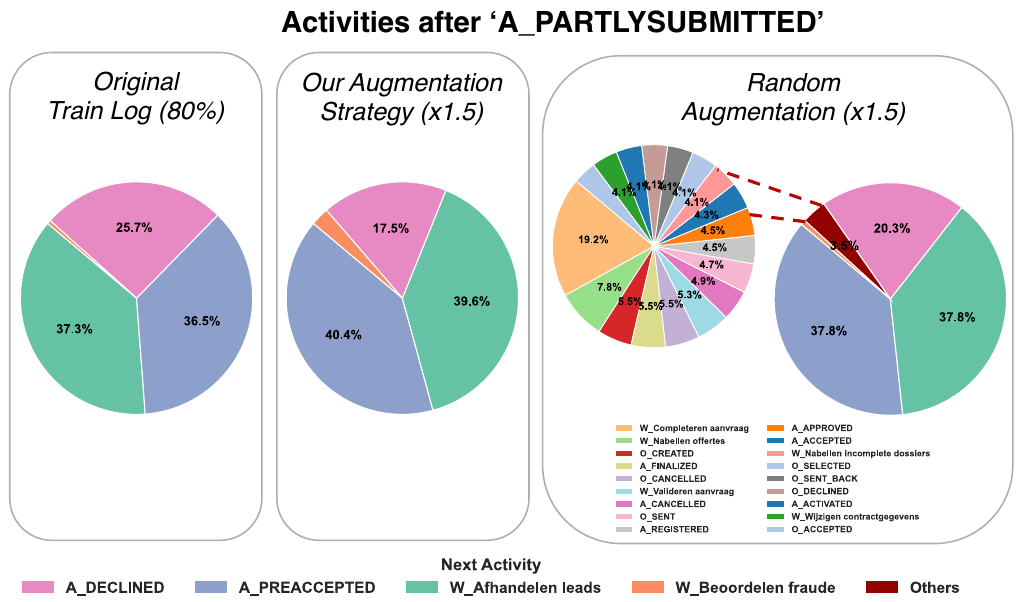}
    \caption[]{Comparison between events directly following \texttt{A\_PARTLYSUBMITTED} in the original train log and the two augmented logs, highlighting the semantic-violating noise injected by the random augmentation method~\cite{kappel2023augmentation} compared to our strategy.}
    \label{fig:split_bpic12}
\end{figure}

\begin{figure}[t]
    \centering
    \includegraphics[width=0.65\linewidth]{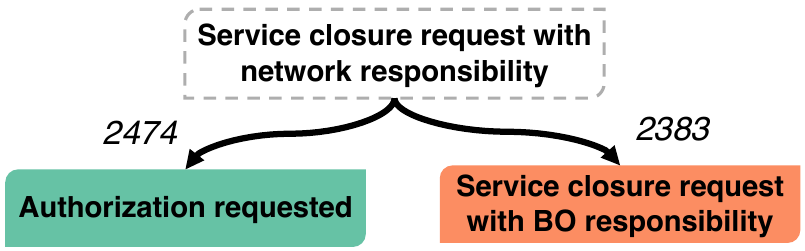}
    \caption{Next activity transitions following the \texttt{Service closure Request with network responsibility} event in the \textit{BAC} dataset. The absolute transition counts of the test set (20\%) establish the ground truth.}
    \label{fig:test_bac}
\end{figure}

\begin{figure}[ht]
    \vspace{-1em}
    \centering
    \includegraphics[width=1\linewidth]{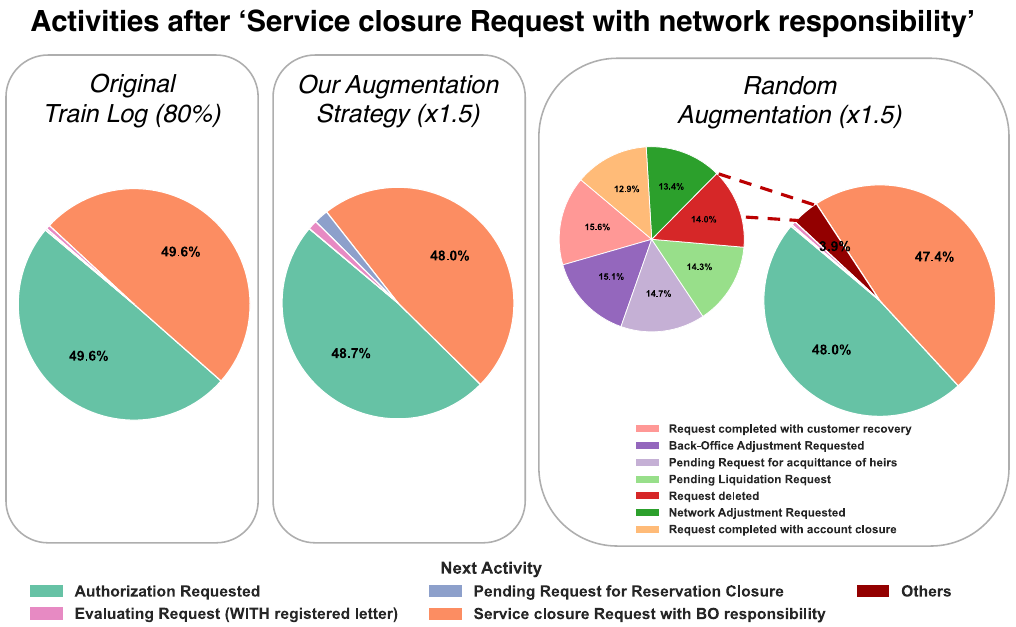}
    \caption[]{Comparison between events directly following \texttt{Service closure Request with network responsibility} in the original train log and the two augmented logs.}
    \label{fig:split_bac}
\end{figure}

For the \textit{BPIC12} dataset, we analyze the events following the \texttt{A\_PARTLYSUBMITTED} event. At this decision point, the control-flow naturally diverges into four valid paths, namely \texttt{W\_Beoordelen fraude}, \texttt{W\_Afhandelen leads}, \texttt{A\_DECLINED} and \texttt{A\_PREACCEPTED}. Fig.~\ref{fig:test_bpic12} establishes the ground truth by displaying the actual transition counts observed in the test set. To evaluate how well the augmentation methods preserve these semantic boundaries, we compare the proportional distributions of the augmented training sets (factor 1.5) in Fig.~\ref{fig:split_bpic12}. The random augmentation method~\cite{kappel2023augmentation} introduces a significant amount of semantic-violating noise. As highlighted in the minority zoom-in of Fig.~\ref{fig:split_bpic12}, it injects a wide range of invalid ``Other" activities into the immediate control-flow. This process-agnostic approach generates trace variants that fundamentally contradict the actual logic of the financial loan application process. Conversely, our proposed augmentation strategy largely mitigates this issue. The results confirm that our generated data remains tightly aligned with the original probability distribution. It still introduces the intended data variability, evidenced by the slight, natural shifts in distribution percentages among the four valid paths, while mitigating the injection of logically impossible activities or arbitrary noise.

We conducted the same analysis on the highly structured \textit{BAC} dataset, isolating events following the event \texttt{Service closure Request with network responsibility}. The test ground truth (cf. Fig.~\ref{fig:test_bac}) favors strict transitions to two primary activities, namely \texttt{Authorization Requested} and \texttt{Service closure Request with BO responsibility}. The strict nature of this process exposes the limitations of random augmentation, which frequently generates semantically invalid ``Other'' events (cf. Fig.~\ref{fig:split_bac}). Our augmentation strategy resolves this issue by leveraging statistically mined patterns. Therefore, it respects the underlying process rules and introduces data variation with limited noise.

\subsection{Running Time} \label{subsec:running_time}
\begin{figure}[h]
    \centering
    \includegraphics[width=1\linewidth]{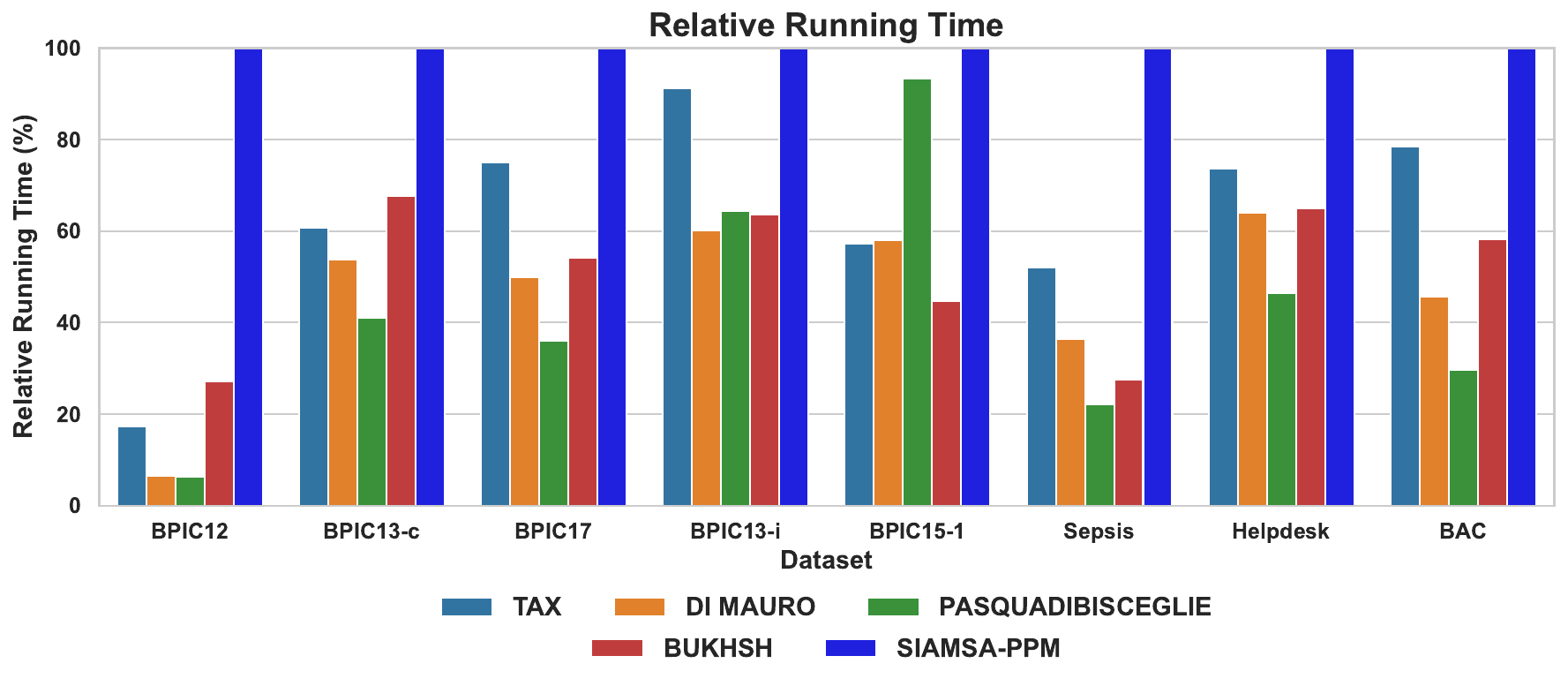}
    \caption{Relative inference time (normalized to the slowest method per dataset) across all benchmark datasets.}
    \label{fig:running_times}
\end{figure}

Fig.~\ref{fig:running_times} reports the relative inference times for all evaluated models, normalized to the slowest method per dataset. As expected, SiamSA-PPM requires more time to make predictions than the baseline methods, especially on complex logs with long average trace lengths like \textit{BPIC12} and \textit{BPIC17}. This added overhead stems directly from model capacity. To capture intricate control-flow dependencies and generalize from augmented data, SiamSA-PPM utilizes a heavy Transformer-based encoder. On the \textit{Helpdesk} dataset, for instance, our architecture contains roughly 666,000 trainable parameters. For comparison, the second-slowest baseline (Tax et al.) uses an LSTM network with only 123,000 parameters. Naturally, a model with more than five times as many parameters demands more extensive matrix operations during the forward pass, which increases running time.

We prioritize predictive performance and robustness over running time. The representational capacity of SiamSA-PPM likely contributes to its performance on data-scarce logs. Since PPM rarely requires strict millisecond latency, the absolute inference times remain highly practical for real-world applications where predictive accuracy is the primary goal.

\subsection{Ablation Study} \label{subsec:ablation_study}
To assess the impact of our augmentation strategy, we conducted a comprehensive ablation study across the four target datasets. As detailed in Table~\ref{tab:accuracy_comparison}, implementing pre-training with the random augmentation strategy yields a modest average improvement over supervised learning (72.56\% versus 71.64\%). Employing our augmentation strategy leads to a further increase to 72.94\%. This gain is not uniform across datasets. On \textit{Sepsis}, supervised training alone (60.87\%) slightly outperforms both pre-training variants, while the largest gains are observed on \textit{BPIC13-c} and \textit{BAC}. Overall, these results indicate the value of our Siamese pre-training method in combination with the statistical transformations.

\begin{figure}[h]
    \centering
    \includegraphics[width=\linewidth]{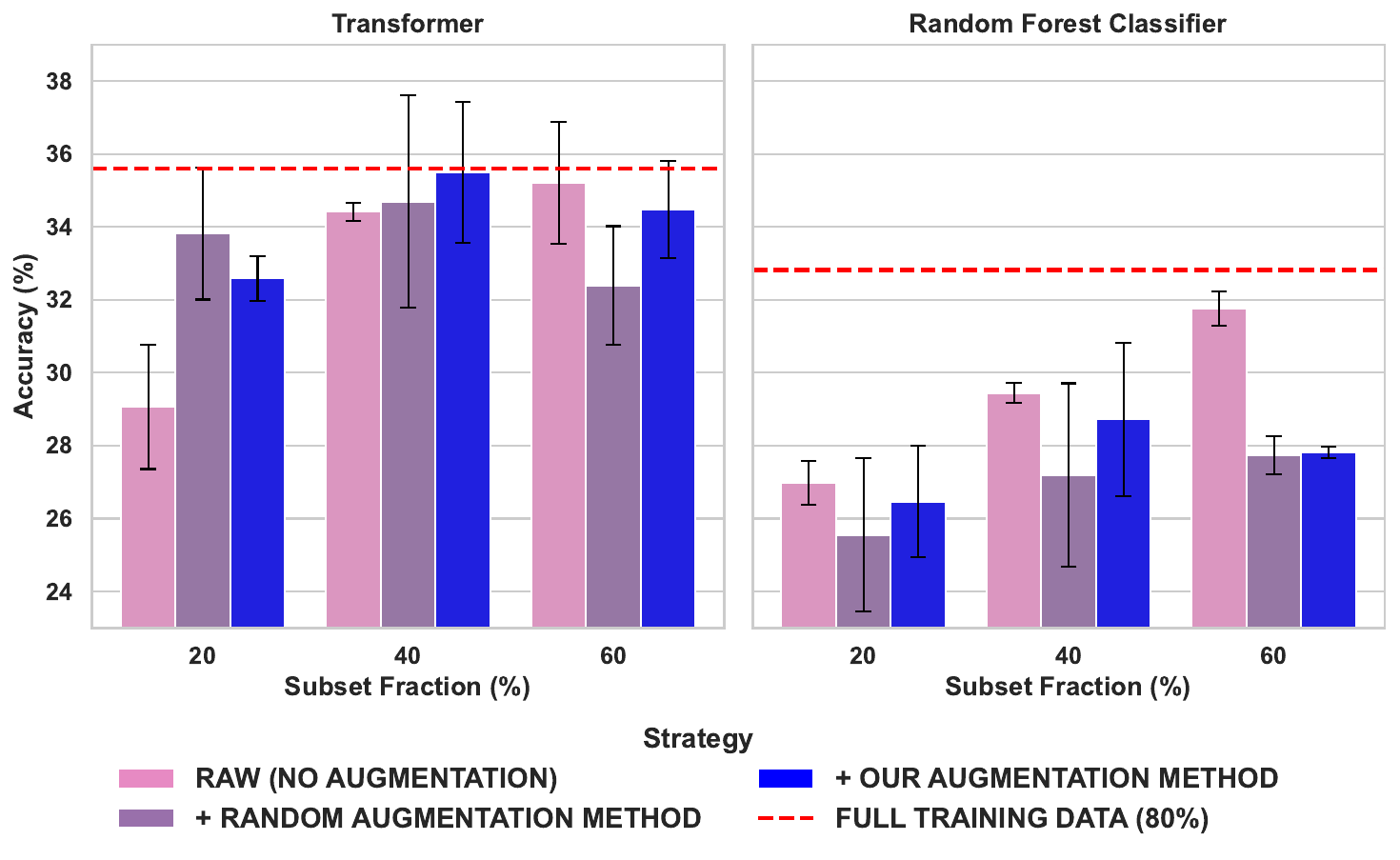}
    \caption{Accuracy comparison between a deep learning Transformer model (left) and shallow Random Forest Classifier (right) on the \textit{BPIC15-1} dataset, under varying raw training subset sizes (20\%, 40\% and 60\%). Each strategy (no augmentation, random augmentation and our augmentation method) uses the same base data, but the latter two are upsampled to the original 80\% training size. The red dashed line denotes the baseline of training on the full 80\% raw training data.}
    \label{fig:ablation_study}
\end{figure}

\begin{table}[ht]
\centering
\caption{Ablation study for the three training strategies over the \textit{target} datasets.}
\label{tab:accuracy_comparison}
\renewcommand{\arraystretch}{1.2}
\resizebox{\textwidth}{!}{
\begin{tabular}{lccccc}
\toprule
\textbf{Training Strategy} & \multicolumn{4}{c}{\textbf{Test Accuracy (\%)}} & \textbf{Average} \\
\cmidrule(lr){2-5} \cmidrule(lr){6-6}
& \textit{Helpdesk} & \textit{BPIC13-c} & \textit{Sepsis} & \textit{BAC} & \\
\midrule
Supervised only                  & $78.50 \pm 0.13$ & $55.67 \pm 0.21$ & $60.87 \pm 0.53$ & $91.50 \pm 0.10$ & 71.64 \\
+ Pre-training (random strategy) & $78.58 \pm 0.23$ & $56.09 \pm 0.28$ & $60.13 \pm 0.90$ & $95.43 \pm 0.06$ & 72.56 \\
+ Pre-training (our strategy)    & $78.60 \pm 0.18$ & $57.06 \pm 0.24$ & $60.62 \pm 0.20$ & $95.49 \pm 0.01$ & $\mathbf{72.94}$ \\
\bottomrule
\end{tabular}%
}
\end{table}
Fig.~\ref{fig:ablation_study} illustrates the impact of data augmentation on model accuracy using varying fractions (20\%, 40\%, 60\%) of the \textit{BPIC15-1} training data. We compare a Transformer model and a Random Forest classifier, each trained on raw subsets, subsets augmented using a random strategy and subsets augmented using our proposed method. For the Transformer, our method improves accuracy at 20\% and 40\% sample sizes, reaching performance close to the full 80\% training data at 40\%, while at 60\% it performs slightly below the unaugmented subset. Interestingly, on the smallest subset (20\%), random augmentation performs slightly better, likely due to the limited information in the smaller number of traces. In contrast, the Random Forest classifier shows no performance gains with either augmentation method, remaining well below the full-data baseline. These results suggest that deep learning models are better equipped to exploit synthetic variation introduced through augmentation, leading to better performance.

\section{Conclusion}\label{sec:conclusion}

\subsection{Discussion} \label{subsec:discussion}
A potential concern with event log augmentation is the validity of the generated traces, particularly for final outcome prediction. Critics might argue that artificially altering the control-flow could change a trace's final outcome, like turning a sequence meant for loan rejection into an approval. However, our statistically grounded approach effectively mitigates this risk by respecting the underlying model constraints. Unlike process-agnostic methods that inject random noise, our statistical transformations adhere to frequent behavioral patterns and XOR-splits mined directly from the original training log. This makes it highly probable that the core control-flow and original outcome label remain consistent, although random augmentors are used as a fallback when no statistical transformation applies.

Furthermore, this augmentation framework may also offer value beyond PPM. Real-world event logs often lack variability due to the limited observation periods or heavy class imbalance, meaning many plausible process variations remain unrecorded. In these scenarios, our strategy helps to ``generalize" the event log by simulating realistic behavior that naturally belongs to the process but was simply missed in the recorded data. This enriched representation provides a more comprehensive baseline of process executions. However, our experiments indicate that not all models benefit equally. While the Transformer model gains from augmentation, the Random Forest classifier does not (cf. Fig.~\ref{fig:ablation_study}). This suggests that exploiting synthetic variation requires models with sufficient representational capacity. Whether augmented logs can benefit other analytical tasks therefore remains an open question for future work.

\subsection{Conclusion} \label{subsec:conclusion}
In this paper, we introduced a novel framework that combines statistically grounded data augmentation techniques with a self-supervised Siamese learning approach to enhance PPM. By addressing the challenges of limited and low-diversity event log data, we demonstrated that our novel transformation methods (StatisticalInsertion, StatisticalDeletion and StatisticalReplacement) preserve process semantics while enriching trace and prefix variability. Coupled with a BYOL-inspired Siamese architecture, SiamSA-PPM successfully learns robust and generalizable representations of process prefixes without relying on manual labels. Extensive experiments across eight real-world event logs confirmed that our method achieves competitive or superior performance in both next activity prediction and final outcome prediction tasks. Finally, our ablation study indicated that pre-training with our augmentation strategy yields a higher average accuracy than pre-training with random transformations.

\subsection{Future Work} \label{subsec:future_work}
Although our results are promising, several avenues remain for future work. To encourage the reader to think about future solutions, we propose several open questions and directions.

\begin{itemize}
    \item \textbf{Addressing skewness in prediction tasks:} Does statistically grounded augmentation make sense for specifically addressing severe skewness in prediction tasks? Future research should investigate whether generating targeted trace variants can be explicitly used for balancing outcome labels in highly imbalanced datasets.
    \item \textbf{Attribute generation:} Can we extend the generation of activity sequences to also accurately synthesize associated resources and timestamps? Expanding the framework to seamlessly handle these complex attributes would significantly enrich the augmented event logs.
    \item \textbf{Enhancing fairness~\cite{deleoni2024fairness}:} How can data augmentation be strategically used to enhance algorithmic fairness in a pretraining fashion? A critical direction involves using the augmentation strategy specifically to balance False Positive (FP) and False Negative (FN) outcomes. Successfully doing so could reduce false positives generated by outcome prediction models, lowering the number of unnecessary interventions and operational costs in Prescriptive Process Analytics (PPA)~\cite{le2025counterfactuals} pipelines.
\end{itemize}

\bibliographystyle{splncs04}
\bibliography{sn-bibliography}

\end{document}